%% file: 0_main.tex
\documentclass[Afour,sageh,times]{sagej}

\usepackage{moreverb,url}
\usepackage{nameref}
\usepackage{color, colortbl}
\usepackage{xcolor}
\usepackage{csquotes}
\usepackage{bigstrut}
\usepackage{multirow}
\usepackage{balance}

\newcommand\BibTeX{{\rmfamily B\kern-.05em \textsc{i\kern-.025em b}\kern-.08em
T\kern-.1667em\lower.7ex\hbox{E}\kern-.125emX}}

\usepackage{hyperref}
 \hypersetup{
     colorlinks=true,
     linkcolor=orange,
     filecolor=orange,
     citecolor=orange,      
     urlcolor=orange,
     }
\usepackage{perpage} %the perpage package
\MakePerPage{footnote} %the perpage package command 
\usepackage{float}
\usepackage{wrapfig}
\usepackage{bbm}
\usepackage{bbold}
\usepackage{algorithm}
\usepackage{algpseudocode}

\newcommand{\p}[1]{\smallskip \noindent \textbf{{#1}.}}
\newcommand{\eq}[1]{Equation~(\ref{eq:#1})}
\newcommand{\fig}[1]{Figure~\ref{fig:#1}}

\begin{document}

% %%%%%%%%%%%%%%%%%%%%%%%%%%%%%%%%%%%%%%%%%%%%%%%%%%%%%%%%%%%%%%%%%%%%%%%%%%%%%%%%%%%%%%%%

\runninghead{Sagheb et al.}

\title{A Unified Framework for Robots that Influence Humans over Long-Term Interaction}

\author{Shahabedin Sagheb\affilnum{1}, Sagar Parekh\affilnum{1}, Ravi Pandya\affilnum{2}, Ye-Ji Mun\affilnum{3}, Katherine Driggs-Campbell\affilnum{3}, Andrea Bajcsy\affilnum{2}, and Dylan P. Losey\affilnum{1}}

\affiliation{\affilnum{1}Virginia Tech, Department of Mechanical Engineering\\
\affilnum{2}Carnegie Melon University, Robotics Institute\\
\affilnum{3}University of Illinois Urbana-Champaign, Department of Electrical and Computer Engineering
}

\corrauth{Dylan P. Losey, 
Department of Mechanical Engineering,
Virginia Tech, 
Blacksburg, VA,
24060, USA.}

\email{losey@vt.edu}

%%%%%%%%%%%%%%%%%%%%%%%%%%%%%%%%%%%%%%%%%%%%%%%%%%%%%%%%%%%%%%%%%%%%%%%%%%%%%%%%%%%%%%%%

\begin{abstract}

Robot actions influence the decisions of nearby humans.
Here \textit{influence} refers to intentional change: robots influence humans when they shift the human's behavior in a way that helps the robot complete its task.
Imagine an autonomous car trying to merge; by proactively nudging into the human's lane, the robot causes human drivers to yield and provide space.
Influence is often necessary for seamless interaction.
However, if influence is left unregulated and uncontrolled, robots will negatively impact the humans around them (e.g., autonomous cars that repeatedly merge in front of humans may cause human drivers to become more aggressive).
Prior works have begun to address this problem by creating a variety of control algorithms that seek to influence humans.
Although these methods are effective in the short-term, they fail to maintain influence over time as the human adapts to the robot's behaviors.
In this paper we therefore present an optimization framework that enables robots to purposely regulate their influence over humans across both \textit{short-term and long-term} interactions.
Here the robot maintains its influence by reasoning over a dynamic human model which captures how the robot's current choices will impact the human's future behavior.
Our resulting framework serves to \textit{unify} current approaches: we demonstrate that state-of-the-art methods are simplifications of our underlying formalism.
Our framework also provides a principled way to \textit{generate} influential policies: in the best case the robot exactly solves our framework to find optimal, influential behavior.
But when solving this optimization problem becomes impractical, designers can introduce their own simplifications to reach tractable approximations.
We experimentally compare our unified framework to state-of-the-art baselines and ablations, and demonstrate across simulations and user studies that this framework is able to successfully influence humans over repeated interactions.
See videos of our experiments here: \url{https://youtu.be/nPekTUfUEbo}
\end{abstract}


\maketitle

%%%%%%%%%%%%%%%%%%%%%%%%%%%%%%%%%%%%%%%%%%%%%%%%%%%%%%%%%%%%%%%%%%%%%%%%%%%%%%%%%%%%%%%%

\input{1_intro}
\input{2_related}
\input{3_problem}
\input{4_current}

\input{5_method}
\input{6_simulations}
\input{7_user-studies}
\input{8_conclusion}

%%%%%%%%%%%%%%%%%%%%%%%%%%%%%%%%%%%%%%%%%%%%%%%%%%%%%%%%%%%%%%%%%%%%%%%%%%%%%%%%%%%%%%%%

\begin{acks}
This research was supported in part by NSF Grants $\#2246446$, $\#2246447$, and $\#2246448$.
\end{acks}

\begin{sm}
\textbf{Code} for our simulations and experiments is available here: \\\url{https://github.com/VT-Collab/influence} 

\medskip 

\noindent \textbf{Videos} of our experiments are available here: \\\url{https://youtu.be/nPekTUfUEbo}
\end{sm}

%%%%%%%% References %%%%%%%%%
\balance
\bibliographystyle{SageH}
\bibliography{references}

\end{document}

%% file: 1_intro.tex
\section{Introduction} \label{sec:intro}

\begin{figure*}[t]
	\begin{center}
        \includegraphics[width=2\columnwidth]{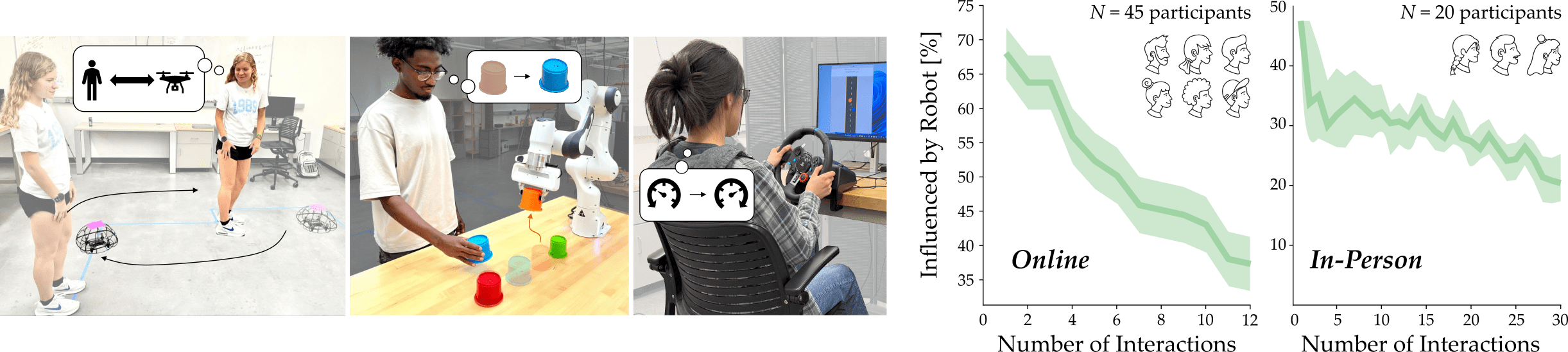}
		\caption{(Left) Human interacting with a drone, a robot arm, and an autonomous car. In each scenario, the robot uses state-of-the-art algorithms to influence the human and change their actions. (Right) Results from online and in-person user studies. The state-of-the-art approaches work in the \textit{short-term}, i.e., human behavior is influenced by the robot in the first few interactions, but not in the \textit{long-term}, i.e., over time the participants adapt to avoid or ignore the influential robot.}
		\label{fig:front}
	\end{center}
    \vspace{-2.0em}
\end{figure*}

When robots and humans interact, it is inevitable that robot behavior will influence human decisions.
Consider an autonomous car moving near a human-driven vehicle.
If the autonomous car drives in an aggressive way --- e.g., merging directly in front of the human --- the human may start to drive more defensively to give the robot space.
Conversely, if the autonomous car is consistently defensive --- e.g., always yielding to the human --- then the human may change their behavior to drive aggressively around the robot.
We refer to these robots as \textit{influential} because they choose actions that i) cause the human's behavior to shift in a way that ii) increases the robot's overall performance.

On one hand, the ability to influence other agents is what enables autonomous robots to seamlessly coordinate with humans. 
On the other hand, if the influential behaviors are haphazardly designed, they can have significant consequences.
In non-embodied domains, we have already observed the effects of influential artificial intelligence (AI) agents on elections, consumer habits, and popular trends \citep{nicas2018youtube, hagen2022rise, stella2018bots,tran2021exploring,khaund2021social}.
Embodied robots are not nearly as pervasive as AI agents; but as robots increasingly integrate into human lives, the scale to which they influence human behaviors will continually grow.
Consider our example of shared roads --- as the number of autonomous vehicles increases, humans have new expectations for how closely they can follow other vehicles, and place more responsibility on autonomous cars for avoiding incidents \citep{schneble2021driver, wired, forbes}.
Prior works have accordingly proposed algorithms to control and regulate influential robots.
Over \textit{short-term} interactions current approaches can discourage or encourage human behaviors \citep{hong2024learning,lazar2018maximizing}, change leader and follower roles within a team \citep{li2021influencing,chari2024optimal}, and manipulate the human's actions to increase the robot's reward \citep{xie2021learning,tian2023towards}.

Despite success in short-term and one-off interactions, robots still struggle to regulate the \textit{long-term} impacts their behaviors have on humans (see \fig{front}).
This stems from a fundamental issue with the way today's robots model human partners.
Prior work on influence typically treats the human as a static agent; i.e., it assumes that the human will always respond to a given robot behavior in a fixed way.
Returning to our motivating example, each time the autonomous car merges in front of the human, it predicts that the human will slow down.
Accordingly --- to influence the speed of the human driver --- the autonomous car simply needs to move into the human's lane.
This approach works well the first few times the human interacts with the robot.
But gradually the human adapts and their responses shift; for instance, after the autonomous car merges into the human's lane several times, the human driver may quickly accelerate to pass the autonomous car or prevent it from merging.

In this paper, we introduce a unifying formalism for influential robots.
Our approach connects prior works that focus on short-term interactions, while also enabling robots to regulate their long-term influence on nearby humans.
To achieve this unification, we recognize that static human models are insufficient: robots must account for how humans continually adapt over time.
Indeed, relying on any single human model is bound to fail --- so we instead propose a control-theoretic framework which accounts for a variety of human responses.
Our core hypothesis is that:
\begin{center}\vspace{-0.3em}
\textit{When robots interact with humans, the robot's current actions will affect not only the human's current response, but also the human's future latent strategy.}
\vspace{-0.3em}
\end{center}
By applying this hypothesis, we are able to formulate influence as an underactuated dynamical system with an unknown and evolving latent state which parameterizes the human's strategy.
The robot is uncertain about how the human's latent strategy will change over time, i.e., how the human will adapt to robot actions.
However, during interaction, the robot gathers information about the dynamic human model, and updates not only its estimate of the current human, but also how that human will respond in the future.
Robots can then leverage this dynamic understanding of the human to select actions that guide the underactuated system towards increased performance.
In practice, applying our control framework results in robots that account for how their current behavior affects the human in the long-term, without relying on a single, pre-specified human model.

Our work is a step towards embodied systems that account for their potentially unintended effects on human decision making.
Overall, we make the following contributions\footnote{Note that a preliminary version of this work was published at the IEEE International Conference on Robotics and Automation \citep{sagheb2023toward}. Results from the preliminary conference paper are included in Section~\ref{sec:E2} and Section~\ref{sec:user1}. However, this journal version is significantly different because i) we develop a unified framework for long-term influence, ii) we demonstrate how prior approaches are approximations of this unified framework, and iii) we conduct new simulations and user studies to demonstrate how the unified framework maintains long-term influence.}:

\p{Formulating Long-Term Influence} 
We present a formal definition for influence in human-robot interaction (\textbf{Section~\ref{sec:problem}}).
When robots interact with humans, the humans' behavior will change in multiple ways --- and not all of these changes are purposeful or meaningful.
We restrict influence to only include robot behaviors that alter the human's strategy in ways that increase the robot's long-term reward.

\p{Optimizing for Influential Behavior}
Leveraging our definition of influence, we derive a dynamical system composed of the robot, human, and environment (\textbf{Section~\ref{sec:unified}}). 
From the robot's perspective, these overall dynamics are an underactuated system, where the immediate and future behavior of the human is indirectly controlled by the robot's actions.
The advantage of this framework is that the robot is not tied to any specific human model; instead, the robot can maintain a distribution of models, and learn how the adaptive human shifts between these models in response to the robot's actions.
More formally, we write the underactuated system as a mixed-observability Markov decision process (MOMDP), where the parameters of the human's short- and long-term dynamics are latent states.
From this MOMDP we can extract an optimal robot policy that maximizes cumulative reward while maintaining long-term influence over the human's latent strategy. 
In practice, our framework enables both precise solutions (for low-dimensional settings) and tractable approximations (for more realistic problems).

\p{Prior Works as Instances of Our Unifying Framework}
Next, we theoretically and empirically demonstrate that current approaches to influence (such as \cite{sadigh2016planning, fisac2019hierarchical, schwarting2019social, xie2021learning, parekh2023learning}) are actually instances of our unifying framework (\textbf{Sections~\ref{sec:existing}}
and \textbf{\ref{sec:U2}}).
This includes robots that influence humans by treating the interaction as a turn-based game, and and robots that learn influential policies by reasoning over latent parameters.
For each of these approaches to influence, we demonstrate that we can reach the same methods by applying different simplifications or approximations to our unified MOMDP framework.
Designers can also apply new approximations to our unifying framework to obtain principled but tractable policies that influence humans in the long-term.

\p{Influencing Humans in the Long-Term}
We perform multiple simulations (\textbf{Section~\ref{sec:sims}}) and user studies (\textbf{Section~\ref{sec:user}}) to explore how our unified approach influences human over repeated interactions.
Our simulations align with our theoretical analysis and suggest that state-of-the-art algorithms are approximations of our unified approach.
Across pursuit-evasion, driving, and manipulation environments, robots that apply our unified framework influence simulated humans more consistently than existing methods.
We then conduct two user studies: in the first, $N=11$ participants interact with an aerial drone, and in the second, $N=20$ participants drive a simulated car alongside an autonomous vehicle.
Across more than $25$ repeated interactions, we find that our unified approach --- and tractable algorithms generated from that approach --- regulate long-term influence on the human more effectively than state-of-the-art baselines.

%% file: 2_related.tex
\section{Related Works} \label{sec:related}

\p{Influence in HRI} As robots enter human environments, influencing humans is inevitable. Robots on factory floors have already been influencing how human workers complete tasks in ways that reduce idle time and increase team efficiency \citep{sanneman2021state, unhelkar2018human, mainprice2013human, pearce2018optimizing, liu2018serocs}. Prior works have explored two fundamental ways in which robots can purposely influence humans: \textit{influence by design} and \textit{influence by action}. Influence by design uses factors such as gender \citep{siegel2009persuasive, bryant2020should}, posture \citep{obaid2016stop}, transparency \citep{wright2019agent, hellstrom2018understandable}, or appearance \citep{breazeal1999build, natarajan2020effects} to alter the human's willingness to collaborate. Although we do not focus on influence by design here, we recognize the importance of social factors for influential robots. In this work, we focus on influence by action. Here, robots select their behavior (e.g., their actions or policy) to guide humans towards advantageous states \citep{newman2020examining, xie2021learning, parekh2022rili, tian2023towards}, increase the team's reward \citep{fisac2019hierarchical, sadigh2016planning, tian2022safety, pandya2024towards}, or change underlying leader and follower roles \citep{reily2020leading, li2021influencing}. For instance, the way a robot arm passes objects to a human can influence how the human grasps and holds these objects \citep{bestick2017implicitly, kedia2024interact}. Robots can also explicitly measure the influence that they have on humans they interact with by computing the divergence between the two agents' future trajectories \citep{tolstaya2021identifying}, seeing if removing the robot changes the human's behavior \citep{hsu2023interpretable, schaefer2021leveraging}, or measuring the deviation from the human's nominal behavior \citep{ding2025surprise}. We instead recognize that not all deviation from nominal behavior is necessarily meaningful, so we formally define influential robot behavior as actions that both change the human's underlying policy and result in \textit{higher reward}. This enables us to explicitly optimize for the \textit{long-term} behavior of robots to have positive influence on humans.

\p{Influence via Games} One way for robots to choose influential actions is to pose and solve a multi-agent game. Prior works have utilized game theoretic approaches to identify control policies that maximize the robot's total reward \citep{nikolaidis2017game, hadfield2016cooperative, fridovich2020efficient, peters2020inference, peters2021inferring, le2021lucidgames, mehr2023maximum, schwarting2019social}. Influential actions arise \textit{naturally} as part of this optimization. This is particularly evident in leader-follower games, known as Stackelberg games, where the leader \textit{acts} (e.g., the robot chooses its actions first) and then the follower \textit{reacts} (e.g., the human selects their response given the robot's chosen behavior). Works such as \cite{sadigh2016planning, lazar2018maximizing, fisac2019hierarchical, schwarting2019social} treat human-robot interaction as a Stackelberg game to solve for influential policies and \cite{li2021influencing, tian2022safety} additionally consider changing the leader and follower roles online. Optimal robots in these Stackelberg games intentionally take actions that maximize the robot's reward by shaping the human's response \citep{ratliff2019perspective}. Other work has used a zero-sum game formulation for learning a robot policy that can influence the human towards safer outcomes \citep{hu2023deception, pandya2025robots}. Outside human-robot interaction, work on multi-agent games with reinforcement learning has shown that it can be beneficial for agents to influence how the other agents will \textit{learn} in the environment \citep{foerster2018learning, letcher2019stable, kim2022influencing, lu2022model} and that including a positive reward for influencing other agents can encourage exploration and cooperation \citep{jaques2019social, yang2020learning}. While all these game-theoretic approaches generate influential actions, they assume both that the robot knows the human's reward function (or it can be estimated \textit{a priori}), and that humans are static agents and will always react the same way. In practice, human behavior shifts over time as they learn from and adapt to the robot's behavior. We accordingly present an optimization framework that can account for such shifts by allowing the robot to \textit{learn} the human's policy and how it evolves online as a function of the robot's current behavior in long-term interactions. 

\p{Influence via Latent Representations} Another line of work has shown success in learning influential robot policies by learning a \textit{latent representation} of the human's strategy or intention. Specifically, some work has learned the dynamics of this latent strategy between interactions \citep{xie2021learning, wang2022influencing, parekh2022rili} or within a single interaction \citep{bajcsy2024learning, li2021influencing}. Other work has learned a latent strategy space offline from a fixed dataset for playing collaborative games using offline reinforcement learning \citep{hong2024learning} and imitation learning \citep{wang2022co}. Similarly, \cite{carroll2024ai} proposes a formalism called ``Dynamic Reward MDPs'' where an AI model may influence a human user's reward function (i.e. their latent strategy) during an interaction. While these approaches do account for how the human's latent intention may change over time, they still assume that the long-term dynamics are stationary --- meaning that they do not capture the full dynamics of the interaction. We show that these latent representation formulations are instances of our unified framework which ignore the long-term dynamics of the human's intentions that depend on the robot's behavior.

\p{Short-Term vs Long-Term Influence} Prior work has been very successful optimizing for influential behavior for robots by predicting how their actions will affect a human's trajectory \citep{tolstaya2021identifying, tang2022interventional, huang2023conditional}, solving games \citep{sadigh2016planning, fisac2019hierarchical}, or learning the dynamics of a latent representation \citep{xie2021learning, bajcsy2024learning}. However, these prior works all focus on \textit{short-term} influence --- while the human may react to the robot (e.g. yielding to an autonomous car), they do not consider how the human's \textit{long-term} behavior will be influenced (e.g. shifting from defensive to aggressive driving). We know from literature in AI that algorithms and recommendation systems have the ability to influence the long-term preferences of humans that interact with them, often to their detriment \citep{nicas2018youtube, hagen2022rise, adomavicius2019hidden, franklin2022recognising, bezou2023shape, carroll2023characterizing}. As embodied agents are starting to be integrated into human life \citep{waymo2024review}, this will inevitably be true for robots as well. Prior work has shown that robots quickly lose their ability to influence humans in a simple repeated driving scenario \citep{cooper2019stackelberg} and that human trust over repeated interactions with robots breaks down \citep{ayub2025continual}. As a result, it is very important to understand and optimize for robot behavior that has positive \textit{long-term influence} on humans. While some have studied repeated Stackelberg games inspired by human-robot interaction \citep{cooper2019stackelberg, zhao2023online, donahue2024impact}, prior work lacks any general framework for modeling how robots can continue to exert both short-term and long-term influence on humans. In this work, we thus formally define long-term influence then introduce a unifying framework that ultimately allows the robot to find optimal, influential behavior across both the short and long term. 

%% file: 3_problem.tex
\section{Problem Setting} \label{sec:problem}

In this section, we formulate the class of settings where robots can influence humans.
For ease of explanation, we will focus on interactions between one robot and one human; however, our formulation can be extended to one robot working alongside multiple people.
We find that influence is not present in all human-robot interactions.
Instead, influence is restricted to settings where the human and robot behaviors are \textit{interdependent}.
Specifically, the robot's performance must depend upon the human's actions, and the human must be willing to change their actions in response to the robot.

\p{Dynamics}
Let $s_\mathcal{R}$ be the robot's state, let $s_\mathcal{H}$ be the human's state, and let $s = (s_\mathcal{R}, s_\mathcal{H}) \in \mathcal{S}$ be the overall system state.
The robot takes actions $a_\mathcal{R} \in \mathcal{A}_\mathcal{R}$ and the human takes actions $a_\mathcal{H} \in \mathcal{A}_\mathcal{H}$.
For instance, within our driving example $s$ contains the heading and position of both cars, $a_\mathcal{R}$ is the steering angle and acceleration of the autonomous vehicle, and $a_\mathcal{H}$ is the steering angle and acceleration of the human's car.
Using $t$ to index the current timestep, the system state transitions according to its dynamics:
\begin{equation} \label{eq:P1}
    s^{t+1} = f(s^t, a_\mathcal{R}^t, a_\mathcal{H}^t)
\end{equation}
We emphasize that both the robot and human actions affect the system state $s$. 
Returning to the driving example, the steering and acceleration of the human and robot change the position and heading of both vehicles.
We will assume that this system state $s$ is fully observable, and that the human and robot can observe each other's previous actions. 
In other words, at timestep $t$ the robot senses $s^t$ and $a_\mathcal{H}^{t-1}$, and the human perceives $s^t$ and $a_\mathcal{R}^{t-1}$.

\p{Rewards} 
Both agents have a task they want to accomplish.
We formulate these tasks using reward functions that map the system state $s$ to scalar values.
The robot's reward function is $r_\mathcal{R}(s, \theta_\mathcal{R})$, and the human's reward function is $r_\mathcal{H}(s, \theta_\mathcal{H})$. 
Here $\theta_\mathcal{R} \in \Theta_\mathcal{R}$ and $\theta_\mathcal{H} \in \Theta_\mathcal{H}$ are the parameters of the robot and human reward functions.
For instance, in our driving example the robot's reward function includes the distance the autonomous vehicle travels (i.e., speed) and the distance between the autonomous vehicle and the human car (i.e., safety).
The parameters $\theta_\mathcal{R}$ determine the relative importance of the speed and safety terms.
More generally, $r_\mathcal{R}(s, \theta_\mathcal{R})$ and $r_\mathcal{H}(s, \theta_\mathcal{H})$ could be neural networks, and $\theta_\mathcal{R}$ and $\theta_\mathcal{H}$ could be the weights of those networks.

Each agent knows their own reward function, but they do not know what the other agent is optimizing for.
Hence, the robot does not know $\theta_\mathcal{H}$ and the human does not know $\theta_\mathcal{R}$.
We define $P_0(\theta_\mathcal{R}, \theta_\mathcal{H})$ as the prior distribution over reward parameters; at the start of the interaction, the robot and human sample their reward parameters from this prior.
Depending on the robot and human rewards, the agents could be collaborating (i.e., sharing a common reward function), competing (i.e., attempting to complete different tasks), or a mix of the two.
We assume that the robot's reward parameters remain constant throughout the interaction.

\p{Interdependence}
To study influence, we explore problem settings where the agents' decisions are interdependent.
More formally, we focus on scenarios where the actions of the robot impact the human's reward, and --- conversely --- the human's actions impact the robot's reward.
This interconnection is captured by the reward functions of both agents: we assume that $r_\mathcal{R}$ and $r_\mathcal{H}$ and have a non-trivial dependence on the overall system state $s = (s_\mathcal{R}, s_\mathcal{H})$.
For example, consider rewards that penalize collisions between the autonomous and human-driven vehicles.
These collisions rely on the position of both vehicles, and thus the rewards depend on $s_\mathcal{R}$ and $s_\mathcal{H}$.
In practice, this restricts our analysis to scenarios where the robot's task performance is affected by how the human updates their state $s_\mathcal{H}$, and vice versa.

\noindent \textbf{Interaction.} 
The human and robot interact over a finite horizon that lasts $H$ timesteps.
In some cases this overall interaction is broken into multiple segments. 
For instance, human drivers will interact with autonomous cars at many intersections, and at each intersection the human driver and autonomous car need to decide who crosses that intersection first.
Within these cases there are $N$ repeated interactions of length $T$, such that $H = N \cdot T$, and the overall, long-term interaction is composed of the repeated segments.

In summary, our problem setting occurs during human-robot interaction when i) both agents' actions affect the overall system dynamics, and ii) the reward of each agent depends on the entire system state.
Many real-world situations fall under this problem setting: consider our motivating example of autonomous and human-driven cars, or an assistive robot working alongside a human user.
These conditions result in interconnected agents whose behaviors affect each other's task performance.
The interplay between the human and robot could play out over a single short-term game, or over long-term and repeated interactions.

\p{Influence}
Within this interdependent problem setting we can formally define influential robot behavior.
Let $a_\mathcal{R}^t = \pi_\mathcal{R}(s^{0:t}, a_\mathcal{R}^{0:t-1}, a_\mathcal{H}^{0:t-1}, \theta_\mathcal{R})$ be a robot policy that maps its history of observations to robot actions $a_\mathcal{R}$, and similarly let $a_\mathcal{H}^t = \pi_\mathcal{H}(s^{0:t}, a_\mathcal{R}^{0:t-1}, a_\mathcal{H}^{0:t-1}, \theta_\mathcal{H})$ be a human policy that maps their history of observations to human actions $a_\mathcal{H}$.
Without loss of generality, the robot begins with some initial policy $\pi_\mathcal{R}^0$, and the human has a baseline policy $\pi_\mathcal{H}^0$.
For instance, within our motivating example the robot's objective is to ensure the human driver maintains a safe speed. 
The autonomous car might initially plan to stay in the right lane, and the human is initially an aggressive driver speeding along the left lane.
We say that the robot policy $\pi_\mathcal{R}'$ is \textit{influential} if i) $\pi_\mathcal{R}'$ has a higher expected cumulative reward than $\pi_\mathcal{R}^0$, and ii) the human follows a policy different from $\pi_\mathcal{H}^0$ when interacting with $\pi_\mathcal{R}'$.
Put another way, robots influence humans when they take actions that increase their own performance while manipulating the human's behavior.
Returning to our example, an influential autonomous car might merge into the human's lane --- causing the human to switch from aggressive to defensive driving while increasing the robot's reward for maintaining safe driving speeds.

%% file: 4_current.tex
\section{Existing Approaches and Long-Term Influence} \label{sec:existing}

Recent works have created learning and control algorithms that regulate influence within our problem setting.
Although these methods can successfully influence humans in the short-term, here we explore how humans respond to the robot over repeated, long-term interactions.
We begin by providing a general formulation for existing approaches (see Section~\ref{sec:E1}).
As we will show, these approaches rely upon a fixed model of the human, i.e., they assume that the human consistently responds to the same robot behaviors in the same way.
We hypothesize that this is a fundamental limitation that will cause unregulated influence in the long-term --- because humans adapt, actions that were once influential may later be avoided or ignored.
To test our hypothesis, we perform online and in-person user studies that measure how effectively today's robots influence humans as a function of time (see Section~\ref{sec:E2}).
Our results suggest that --- although existing frameworks can correctly influence human's over a handful of interactions --- these approaches fail as humans adapt to the robot's behaviors.

\subsection{Formulating Existing Methods} \label{sec:E1}

We first describe how robots currently choose influential actions.
Existing approaches are based on optimization: the robot seeks to maximize its cumulative reward, while recognizing that both the robot's and human's behaviors contribute to that interdependent reward.
The robot \textit{directly} controls its own behaviors.
In order to \textit{indirectly} influence the human's behaviors, the robot is equipped with a human model.
This model --- which can be learned or pre-programmed --- predicts how the human will react to the robot's actions.
For instance, within our driving example this human model might predict that the human will yield if the autonomous car merges in front of the human driver, and accelerate otherwise.
Robots then reason over the human model to select actions (e.g., merging in front of the human) that will manipulate the user and optimize the robot's overall reward (e.g., minimizing the human driver's speed).

\p{Trajectories} 
We formulate the underlying optimization over a finite horizon of $H$ timesteps.
Let the sequence of robot actions over $H$ timesteps be $\mathbf{a}_{\mathcal{R}} = (a_{\mathcal{R}}^0, \hdots, a_{\mathcal{R}}^{H})$, and similarly let $\mathbf{a}_{\mathcal{H}} = (a_{\mathcal{H}}^0, \hdots, a_{\mathcal{H}}^{H})$ be the human's action sequence.
If we execute these action sequences starting at state $s^0$ and following the dynamics from \eq{P1}, we obtain a trajectory of system states.
This trajectory can be written as: $\xi(s^0, \mathbf{a}_{\mathcal{R}}, \mathbf{a}_{\mathcal{H}}) = (s^0, \ldots, s^{H+1})$.
Note that the robot directly controls $\mathbf{a}_{\mathcal{R}}$, but the human's action sequence $\mathbf{a}_{\mathcal{H}}$ relies on the person the robot is interacting with.

\p{Total Reward}
The robot seeks to maximize its cumulative reward across the interaction.
Given a trajectory $\xi$, the robot's total reward is the sum of state rewards:
\begin{equation} \label{eq:E1}
    R_{\mathcal{R}}(s^0, \mathbf{a}_{\mathcal{R}}, \mathbf{a}_{\mathcal{H}}) = \sum_{s \in \xi(s^0, \mathbf{a}_{\mathcal{R}}, \mathbf{a}_{\mathcal{H}})} r_\mathcal{R}(s, \theta_\mathcal{R})
\end{equation}
Likewise, the human's cumulative reward is the sum of state rewards $r_\mathcal{H}(s, \theta_\mathcal{H})$ along the trajectory $\xi$.
Because these rewards encode the desired tasks (e.g., ensuring the human driver maintains a safe speed), maximizing total reward corresponds to efficient task performance.

\p{Optimization}
Existing approaches achieve influential behaviors by maximizing \eq{E1} in interdependent problem settings.
We can generally break down these approaches into two components: i) an optimizer that seeks to maximize the robot's cumulative reward, and ii) a model that predicts how the human will react to robot actions.
Both components are formulated below:
\begin{equation} \label{eq:E2}
    \begin{aligned}
            \mathbf{a}_\mathcal{R}^* &= \text{arg}\max_{\mathbf{a}_\mathcal{R}}~R_\mathcal{R}\big(s^0, \mathbf{a}_\mathcal{R}, g(s^0, \mathbf{a}_\mathcal{R})\big) \\
            & \text{s.t.} \quad s^{t+1} = f(s^t, a_\mathcal{R}^t, a_\mathcal{H}^t)
    \end{aligned}
\end{equation}
where $\mathbf{a}_\mathcal{R}^*$ is the optimal sequence of $H$ robot actions, and $\mathbf{a}_\mathcal{H} = g(s^0, \mathbf{a}_\mathcal{R})$ is the human model.
Influential behaviors naturally emerge by solving \eq{E2} because the robot considers how its own actions guide the human's response.
If the robot can identify behaviors that cause an advantageous human reaction, the robot will select those behaviors and attempt to influence the human.
Take our running example where the autonomous car seeks to regulate the speed of a human driver: the human model predicts that merging into the human's lane will slow that driver.
The robot therefore performs a merge --- and influences the human to slow down --- in order to maximize its overall reward.
In theory, \eq{E2} can extend across an arbitrary time horizon $H$.
In other words, this framework could find behaviors that influence humans in the short-term and in the long-term.

\p{Human Models}
Existing works generally agree on the framework in \eq{E2}.
However, these approaches differ in their choice of the human model; i.e., how they instantiate the function $g$.
Below we summarize two common themes from the literature.

Methods such as \cite{sadigh2016planning, lazar2018maximizing, fisac2019hierarchical, schwarting2019social, losey2019robots, fridovich2020efficient, tian2022safety} model the human as a game-theoretic agent that seeks to maximize their own reward.
If the human and robot act simultaneously, the human model solves for the Nash Equilibrium; alternatively, if the human and robot take turns, this can be treated as a Stackelberg game.
In either case, here the human model becomes:
\begin{equation} \label{eq:E3}
    \begin{aligned}
            g(s^0, \mathbf{a}_\mathcal{R}) &= \text{arg}\max_{\mathbf{a}_\mathcal{H}}~R_\mathcal{H} (s^0, \mathbf{a}_\mathcal{R}, \mathbf{a}_\mathcal{H}) \\
            & \text{s.t.} \quad s^{t+1} = f(s^t, a_\mathcal{R}^t, a_\mathcal{H}^t)
    \end{aligned}
\end{equation}
where $R_\mathcal{H}$ is the human's cumulative reward.
Existing methods assume that the robot knows the human's reward function, or the robot can learn that reward function from prior data.
Combining \eq{E3} with \eq{E2} results in nested optimization, where the robot reasons over how the optimal human will respond to its actions $\mathbf{a}_\mathcal{R}$.

Alternatively, methods such as \cite{xie2021learning, bajcsy2024learning, jaques2019social, li2021influencing, chen2020trust} model the human as an agent whose decisions are parameterized by some latent variable $z$.
Here the human's policy
$\pi_\mathcal{H}(s, z) \rightarrow a_\mathcal{H}$ maps system states and the latent $z$ to human actions.
This latent parameter could be the human's willingness to adapt, the human's strategy for completing the task, or the human's reward or preferences when interacting with the robot.
Importantly, $z$ updates over time in response to the robot's behavior.
Consider a setting where the human and robot repeatedly interact $N$ times, and let $i$ index the current interaction.
In this case the human model from \eq{E2} becomes:
\begin{equation} \label{eq:E4}
    \begin{aligned}
        g(s^0, \mathbf{a}_\mathcal{R}) &= \pi(s, z^i), \quad i = 1, \ldots, N \\
        & \text{s.t.} \quad z^{i} = g(z^0, s^0, a_\mathcal{R}^{0:i})
    \end{aligned}    
\end{equation}
where $z^{i} = g(z^0, s^0, a_\mathcal{R}^{0:i})$ captures how the latent parameter updates between interactions, and $a_\mathcal{R}^{0:i}$ includes all robot actions up to the start of interaction $i$.
In practice, $z$ could express whether the human is an aggressive or defensive driver.
If the human is initially aggressive, and the autonomous car merges into its lane, the human may become wary of the autonomous car and switch to more defensive driving.
Under this model the robot influences the human by guiding $z$ towards values that align with the robot's underlying objective (e.g., manipulating the human to be a defensive driver so that the human slows down).

\subsection{Testing Long-Term Influence} \label{sec:E2}

\begin{figure*}[t]
	\begin{center}
        \includegraphics[width=2\columnwidth]{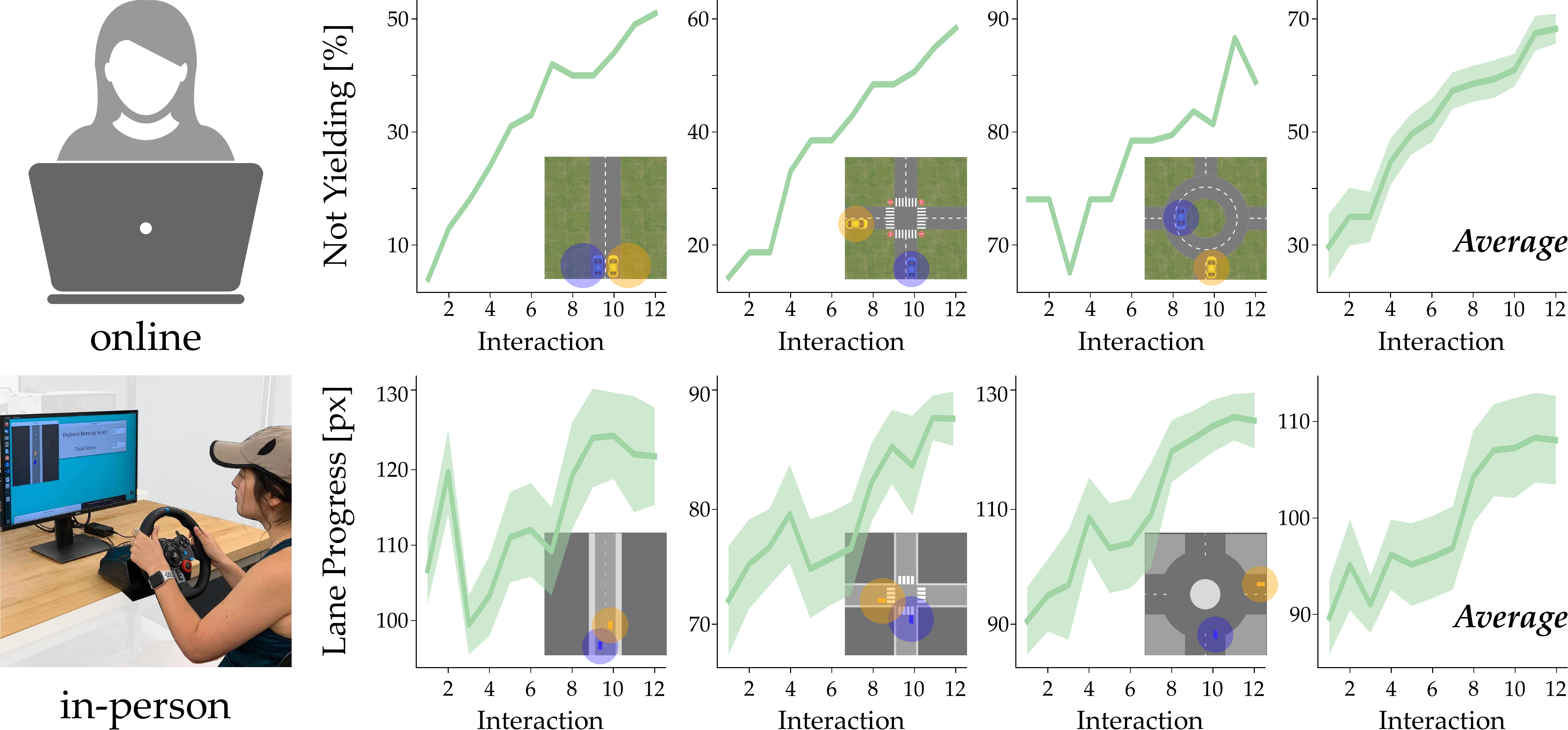}
		\caption{Exploring how the existing approaches from Section~\ref{sec:E1} influence humans over short- and long-term interactions.
        Participants repeatedly interact with an autonomous car that uses \eq{E2} and the static human model from \eq{E3} to influence their behavior. 
        The autonomous car selects actions $\mathbf{a}_{\mathcal{R}}$ by treating each interaction as an independent Stackelberg game; this is consistent with prior works \citep{sadigh2016planning, fisac2019hierarchical, tian2022safety, schwarting2019social}. The robot is rewarded for influencing the human to slow down, yield, and reduce lane progress. For both \textbf{online} (Top) and \textbf{in-person} (Bottom) participants, \textit{the robot's influence decreases over time} (i.e., human yields less or makes more lane progress). In the last column (Right) we display the average behavior across all three environments. Shaded regions show standard error.}
		\label{fig:us1}
	\end{center}
    \vspace{-2.0em}
\end{figure*}

In Section~\ref{sec:E1}, we outlined the general optimization framework existing robots use to influence human actions.
At a high-level, this framework seeks to maximize the robot's reward, and at a low-level, the robot achieves reward maximization by reasoning over a human model.
But the two types of human models currently employed are \textit{static}.
Both \eq{E3} and \eq{E4} state that the human will change behaviors in response to the robot's actions.
However, this change follows a consistent pattern; e.g., if the autonomous car merges, the human driver slows down.
In practice, real humans adopt and change --- making robot behaviors that were once influential now ineffective.

In Section~\ref{sec:E2}, we now explore this limitation by measuring how an existing method influences humans over \textit{long-term} interaction.
Across online and in-person user studies, participants drive a simulated vehicle while sharing the road with an autonomous car.
The autonomous car selects actions according to \eq{E2} while using the static human model in \eq{E3} (i.e., the robot applies the game-theoretic method from \cite{sadigh2016planning, fisac2019hierarchical, tian2022safety, schwarting2019social}). 
Each participant interacts with the autonomous car across highway, intersection, and roundabout  environments for $36$ total trials. 
Our results from $45$ online users and $10$ in-person drivers show that the robot is initially able to influence humans to yield, but over time human drivers adapt to ignore or avoid the autonomous car.

\p{Experimental Setup} 
Participants shared the road with an autonomous car in three settings: highway, intersection, and roundabout (see \fig{us1}). 
These settings are consistent with related works \citep{sadigh2016planning, tian2022safety, fisac2019hierarchical, fridovich2020efficient, schwarting2019social}. 
To simulate the driving environment and vehicle dynamics in real-time we used CARLO \citep{cao2020reinforcement}. 
\textit{In-person} participants controlled their simulated vehicle using a Logitech G29 steering wheel and responsive pedals. 
Each interaction ended after a fixed number of timesteps. 
\textit{Online} participants first watched an animated video of the start of the interaction, and then selected their behavior from a multiple choice menu (i.e., participants could choose to yield or try and pass the autonomous car). 
Both in-person and online participants earned points for avoiding a collision, staying on the road, and making lane progress. 
We displayed the participant's current score throughout the experiment. 
All participants interacted within the highway, intersection, and roundabout settings $12$ times each for a total of $36$ repeated interactions. 
The presentation order for each road setting was randomized and balanced across users.

\p{Independent Variables} 
The autonomous car solved the optimization problem in \eq{E2} and \eq{E3} to select its actions $\mathbf{u}_{\mathcal{R}}$. 
More specifically, the robot treated each interaction as a separate Stackelberg game where the autonomous car played first and the human model predicted the optimal response to the robot's actions.
We rewarded the robot for avoiding collisions and minimizing the human's lane progress (i.e., slowing down the human driver). 
More specifically, we selected:
\begin{equation} \label{eq:U1}
    r_\mathcal{R}(s, \theta_\mathcal{R}) = - \dot{s}_\mathcal{H} - 10 \cdot \mathbbm{1}\{\text{collision in }s\}
\end{equation}
where $\dot{s}_\mathcal{H} \subset s$ is the velocity of the participant's vehicle. 
The robot assumed that the human's reward matched the score that was displayed on the screen:
\begin{multline} \label{eq:U2}
    r_\mathcal{H}(s, \theta_\mathcal{H}) = \dot{s}_\mathcal{H} - 10 \cdot \mathbbm{1}\{\text{on road in }s\} \\ -  100 \cdot \mathbbm{1}\{\text{collision in }s\}
\end{multline}
Positive values for $\dot{s}_\mathcal{H}$ indicate that the human's car is moving forward along the road (i.e., making lane progress), while negative values mean the human's car is in reverse. 
In order to maximize the state reward from \eq{U1}, the robot attempts to influence humans to yield, slow down, or reduce their overall lane progress.

\p{Dependent Variables} 
For \textit{online} participants, we recorded whether the human chose to yield or pass the autonomous car. 
For \textit{in-person} subjects we measured their lane progress, i.e., the vertical distance they traveled.
In each environment, the human's car started at the bottom of the screen and drove towards the top of the screen; a driver that never yields to the autonomous car would maximize their lane progress.

\p{Participants} 
For the \textit{online} component of the user study we recruited $63$ anonymous participants. 
At the start of the experiment these participants read our instructions and then answered qualifying questions to check that they understood the experimental procedure. 
A total of $45$ users passed these questions and continued on to the survey.

For the \textit{in-person} component we recruited $12$ participants from the Virginia Tech community. 
Of these, $10$ answered the qualifying questions correctly and completed the experiment ($5$ female, ages $24.7 \pm 5.2$ years). 
All participants provided informed written consent consistent with university guidelines (IRB \#$20$-$755$). 
We recognize that users may adapt to become better drivers as they continue to interact in our simulated environment. 
To account for this confounding factor we had participants practice driving without any autonomous cars in each environment until they were able to consistently reach expert-level scores.

\p{Hypothesis} We hypothesized that:
\begin{quote}
\p{H1} \textit{As the number of interactions increases, existing approaches will fail to regulate how they influence the human driver.}
\end{quote}

\p{Results} 
Our results from this initial user study are summarized in \fig{us1}. 
The top row shows how frequently the online users chose not to yield to the autonomous car as a function of interaction number; the bottom row displays lane progress for in-person drivers over repeated interactions. 

\textit{Online} users chose to either yield or pass the autonomous car during each interaction. 
We performed Wilcoxon signed-rank tests to see how the human's choice evolved between the first interaction and the final interaction (i.e., to see how the human's response to the robot changed over time). 
Our results averaged across all three driving scenarios reveal that humans passed the autonomous car more frequently by the end of experiment ($Z = -5.798$, $p < .001$). 
This change was also statistically significant for the highway ($Z = -4.583$, $p < .001$) and intersection environments ($Z = -3.838$, $p < .001$), 
but not for the roundabout environment ($Z = -1.155$, $p = .248$). 
Within the roundabout, humans rarely yielded to the robot, perhaps because they already perceived their own car as having the right of way.

For \textit{in-person} drivers we measured their lane progress. 
Remember that the autonomous car is trying to influence humans to reduce their speed; as such, higher lane progress is correlated with less robot influence. 
Paired t-tests show the human's average lane progress was significantly higher at the final interaction as compared to their first interaction ($t(29) = -5.952$, $p < 0.001$). 
This trend is consistent across highway ($t(9) = -2.4, p < .05$), intersection ($t(9) = -3.1, p < .05$), and roundabout ($t(9) = -5.6, p < .001$).
The result suggests that robots were able to influence humans to slow at the start of the experiment, but as the number of interactions increased the robot was no longer able to influence the human effectively.

\p{Summary}
Our results from this study support \textbf{H1}. 
Autonomous cars that leverage an existing game-theoretic framework to generate influential behaviors are effective in the \textit{short-term}, but do not maintain the same influence across the \textit{long-term}.
Over repeated interactions, human drivers adapt to ignore the robot's influential actions (in the online study) or anticipate the robot and avoid its behavior (in the in-person study).
This suggests that existing approaches to influence are insufficient; if we want robots that can regulate how they manipulate the human's actions in the long term, we need to re-examine our underlying framework.

%% file: 5_method.tex
\section{A Unified Framework for Long-Term Influence} 
\label{sec:unified}

Existing robot policies fail to maintain long-term influence during human interactions.
Based on our experiments, we hypothesize that this failure occurs because existing methods rely on \textit{static} human models.
Under static models, the robot assumes the human has a fixed strategy for reacting to the robot's actions.
But --- as we observed in Section~\ref{sec:E2} --- the human's response often evolves over time (e.g., accelerating to prevent the autonomous car from merging).
In Section~\ref{sec:U1}, we now capture dynamic humans by reasoning over history-aware human models.
These models predict both short-term human behaviors (e.g., yielding to the autonomous car) as well as long-term shifts (e.g., the human adapting from defensive to aggressive driving).
Optimizing over \textit{dynamic} human models leads to our unified framework for short- and long-term influence.
We ultimately formulate influence as a mixed-observability Markov decision process (MOMDP).
As shown in Section~\ref{sec:U2}, casting the problem of long-term influence as a MOMDP enables two things.
First, we can compute robot policies that consistently influence the human while accounting for how their current behaviors affect the user in later interactions.
Second, we can demonstrate that existing approaches are actually simplifications of our framework, revealing why these methods fail to control influence across long-term interactions.

\subsection{Optimizing over Dynamic Human Models} \label{sec:U1}

\begin{figure*}[t]
	\begin{center}
        \includegraphics[width=2\columnwidth]{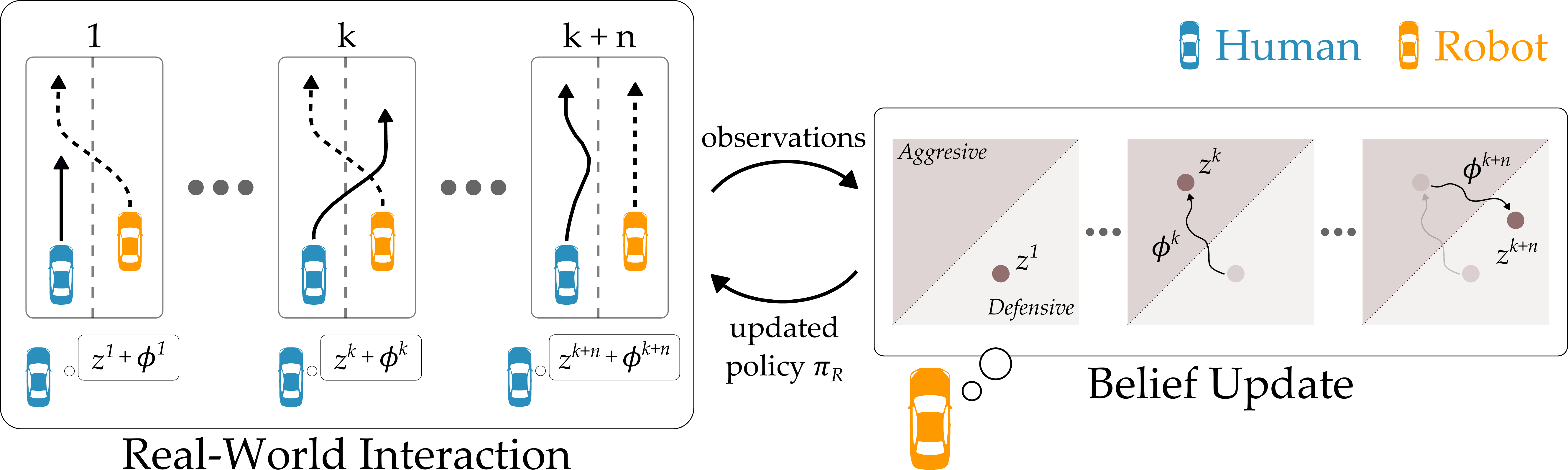}
		\caption{Our unified framework for influence. When interacting with humans, the robot models the human's short-term as well as long-term dynamics. During interaction $1$, the robot merges into the human's lane to influence them. Upon observing the human slow down, the robot models their short term dynamics and infers that the human is a defensive driver $z^1$. However, over $k$ repeated interactions the human might change their response strategy and start driving aggressively according to their set of rules $\phi^k$. Modeling the human's long-term dynamics enables the robot to anticipate the change in the human's response strategy. This enables the robot to optimize for a policy which can influence the human over long-term interaction. During interaction $k + n$, the robot does not merge into the human's lane, anticipating how the human will change lanes to try and avoid the autonomous car.}
		\label{fig:unified}
	\end{center}
    \vspace{-2.0em}
\end{figure*}

Here we introduce our unified framework for long-term influence.
Similar to existing methods, our approach is based on the high-level goal of maximizing the robot's cumulative reward.
However, the key difference is how we model the human agent --- particularly in the long-term.
An overview of our unified approach is shown in \fig{unified}.

\p{History-Aware Human Model} 
Let $\pi_\mathcal{H}^*$ denote the human's actual policy.
By definition, this policy maps some subset of all available human observations to human actions $a_\mathcal{H} \in \mathcal{A}_\mathcal{H}$. 
At timestep $t$ the human can reason over $s^{0:t}$ (the system states up to and including the current state), $a_\mathcal{R}^{0:t-1}$ and $a_\mathcal{H}^{0:t-1}$ (the previous robot and human actions), and $\theta_\mathcal{H}$ (the human's own reward parameters).
Accordingly, the human's true policy is of the form: 
\begin{equation} \label{eq:I1}
    a_\mathcal{H}^t = \pi_\mathcal{H}^*(s^{0:t}, a_\mathcal{R}^{0:t-1}, a_\mathcal{H}^{0:t-1}, \theta_\mathcal{H})
\end{equation}
Intuitively, \eq{I1} means that the human's decisions are based on more than the robot's most recent actions --- the human may reason over the history of interactions between themselves, the robot, and the environment.
This implies that the robot's current behaviors could affect downstream human responses (e.g., the robot actions at interaction $i$ could affect the human's decisions at interaction $i + j$).

\p{Dynamic Human Model} 
The robot does not have access to $\pi_\mathcal{H}^*$.
Instead, the robot maintains its own human model.
We have already seen examples of fixed human models in \eq{E3} and \eq{E4}.
Here we extend these models to capture the history-aware aspects of the human's time-varying decision making.
We rewrite \eq{I1} into:
\begin{align}
    a_\mathcal{H}^t &= \pi_\mathcal{H}(s^t, z^t) \label{eq:I2} \\ 
    z^{t+1} &= g_s(s^t, a_\mathcal{R}^t, a_\mathcal{H}^t, z^t, \phi^t) \label{eq:I3} \\
    \phi^{t+1} &= g_l(s^t, a_\mathcal{R}^t, a_\mathcal{H}^t, \phi^t) \label{eq:I4}
\end{align}
where $\pi_\mathcal{H}$ is the robot's current estimate of the human's policy.
This policy is parameterized by the latent representation $z \in \mathbb{R}^d$.
For instance, within our motivating example $z$ could capture whether the human is an aggressive or defensive driver.
The latent state $z$ updates at two levels: with \textit{short-term} dynamics $g_s$ and \textit{long-term} dynamics $g_l$, with the latent vector $\phi\in\mathbb{R}^k$ parameterizing how the human chooses their next latent strategy $z^{t+1}$.

Imagine our autonomous car on the highway with a human-driven vehicle.
If the autonomous car merges in front of the human at timestep (or interaction) $t$, that could cause the human to slow and drive defensively at timestep (or interaction) $t+1$.
This immediate response is formulated by short-term dynamics $g_s$, which expresses how the human reacts to the robot.
Here human's reaction strategy for selecting $z^{t+1}$ is parameterized by vector $\phi \in \mathbb{R}^k$.
In practice, $\phi$ could be the weights of a neural network $g_s$, or the set of rules which govern the human's response strategy (e.g., slow down if the autonomous car merges, accelerate otherwise).
Prior works in Section~\ref{sec:existing} stop at this point: they leave $\phi$ as a constant value.
But --- to capture how humans adapt over time --- we propose the long-term dynamics $g_l$ that update $\phi$ in \eq{I4}.
Returning to our example, if the autonomous car repeatedly merges in front of the human, at some timestep (or interaction) $\tau$ the human may shift their response strategy.
Instead of yielding to the robot with their original rules $\phi^0$, now the human uses a new set of rules $\phi^\tau$ to react to the robot and update $z$.
The long-term dynamics over $\phi$ enable the robot to model how its current actions could affect the human's downstream behavior (e.g., causing the human to shift from defensive to aggressive driving).

\p{Dynamical System: State and Human Evolution}
Now that we have a human model with short-term and long-term components, we next will integrate this model within a larger representation of the system.
From the robot's perspective, its actions $a_\mathcal{R}$ affect the interaction in two ways.
The robot's behavior causes the environment state $s$ to transition according to the state dynamics from \eq{P1}.
In addition, the robot's actions cause the human's state with parameters $z$ and $\phi$ to evolve according to Equations~(\ref{eq:I3}) and (\ref{eq:I4}).
To capture the robot's overall effect on the environment and human we therefore introduce the \textit{augmented system state} $x = (s, z, \phi)$.
This augmented state has dynamics:
\renewcommand*{\arraystretch}{1.5}
\begin{align}\label{eq:I5}
    x^{t+1} &= \begin{bmatrix} s^{t+1} \\ z^{t+1} \\ \phi^{t+1} \end{bmatrix} = 
    \begin{bmatrix}  
    f\big(s^t, a_\mathcal{R}^t, \pi_\mathcal{H}(s^t, z^t) \big)
    \\ g_s\big(s^t, a_\mathcal{R}^t, \pi_\mathcal{H}(s^t, z^t), z^t, \phi^t
    \big)
    \\ g_l\big(s^t, a_\mathcal{R}^t, \pi_\mathcal{H}(s^t, z^t), \phi^t
    \big) 
    \end{bmatrix} \\ 
    &= 
    \begin{bmatrix} 
    F(x^t, a_\mathcal{R}^t)
    \\ G_S(x^t, a_\mathcal{R}^t)
    \\ G_L(x^t, a_\mathcal{R}^t)
    \end{bmatrix} \label{eq:I6}
\end{align}
Note that in the dynamical system above we have substituted the robot's estimated human policy $\pi_\mathcal{H}(s, z)$ in place of the human actions $a_\mathcal{H}$.
Incorporating this human model results in the augmented dynamics $F$, $G_S$, and $G_L$, which only depend on the augmented state $x$ and the robot's action $a_\mathcal{R}$.

The resulting dynamics in \eq{I6} formulate how the robot can influence the human.
By taking actions $a_\mathcal{R}$ the robot not only i) changes the physical state $s$, but the robot also ii) impacts the human's short-term and long-term latent states $z$ and $\phi$.
If the robot had access to \eq{I6}, it could use these dynamics to select actions that would drive the human towards a beneficial latent state (e.g., causing the human to remain as a safe, defensive driver).

\p{MOMDP}
The augmented state $x$ has mixed observability: the robot can measure $s$, but the human's latent states $z$ and $\phi$ are unknown.
Despite this uncertainty, the robot seeks to maximize its expected cumulative reward subject to the overall, single-agent dynamics presented in \eq{I6}.
We can formulate this optimization problem as an instance of a mixed-observability Markov decision process (MOMDP) \citep{ong2010planning}.
The key advantage of this formulation is that --- if we can solve the MOMDP --- we obtain the optimal robot policy that maximizes expected reward while reasoning over the short-term and long-term human model.

Our influence MOMDP is defined by the tuple $\mathcal{M} = \langle \mathcal{X}, \mathcal{A}_\mathcal{R}, (F, G_S, G_L), \mathcal{O}, r_\mathcal{R}, H \rangle$.
Here $x \in \mathcal{X}$ is the augmented system state, where $s$ is observed by the robot and $z$ and $\phi$ are unknown and must be inferred during the interaction.
The robot takes actions $a_\mathcal{R} \in \mathcal{A}_\mathcal{R}$.
These actions cause the augmented state to transition according to the underactuated dynamics $F$, $G_S$, and $G_L$ from \eq{I6}.
At each timestep the robot makes observation $o^t = (s^t, a_\mathcal{H}^{t-1}) \in \mathcal{O}$ with two parts: the robot measures the current system state $s^t$ and the previous human action $a_\mathcal{H}^{t-1}$.
The robot then uses these observations to update its probability distribution over the human's latent parameters $z$ and $\phi$.
We will refer to this probability distribution as the robot's \textit{belief}.
The robot updates its belief using the human model, where $a_\mathcal{H}^t = \pi_\mathcal{H}(x^t, z^t)$ is the likelihood of action $a_\mathcal{H}^t$ given representation $z^t$, and $z^{t+1} = g_s(s^t, a_\mathcal{R}^t, a_\mathcal{H}^t, z^t, \phi^t)$ is the likelihood of representation $z$ given representation dynamics $g_s$ with parameters $\phi$.
Hence, the human model presented in Equations~(\ref{eq:I2})--(\ref{eq:I4}) provides the observation model within the MOMDP.
The final two components of the MOMDP tuple are the same as our original problem statement.
The robot is trying to optimize its reward function $r_\mathcal{R}(x, \theta_\mathcal{R}) = r_\mathcal{R}(s, \theta_\mathcal{R})$, and the interaction lasts a total of $H$ timesteps.

\subsection{Existing Approaches are Approximations}
\label{sec:U2}

In Section~\ref{sec:U1} we introduced a general human model with short- and long-term dynamics, and then incorporated this human model within the augmented system dynamics.
From the robot's perspective, the resulting dynamics convert the optimization problem into an instance of an MOMDP.
Formulating our problem as a MOMDP enables us to apply existing tools to identify the optimal robot policy $\pi_\mathcal{R}$.

\p{Solving for a Long-Term Influential Policy}
A variety of online solvers have been developed to obtain near-exact estimates of $\pi_\mathcal{R}$.
For example, methods such as POMCP \citep{silver2010monte}, DESPOT \citep{somani2013despot}, and their variants \citep{sunberg2018online, kurniawati2022partially} input the MOMDP $\mathcal{M}$, the current system state $s$, and the current belief over $z$ and $\phi$, and then output the optimal robot action $a_\mathcal{R}$.
Because our MOMDP includes both a short-term and long-term human model, the actions output by these solvers estimate not just how they will immediately influence the human, but also how they will guide the human's internal state for future interactions.
We apply these solvers in Sections~\ref{sec:sims} and \ref{sec:user} to compute influential robot's policies.
Our results empirically demonstrate how the robot policies obtained via this MOMDP maintain influential robot control over humans, maximizing robot reward while still enabling effective coordination.

\p{Generating Tractable Approximations}
The underlying MOMDP structure developed in Section~\ref{sec:U1} provides the \textit{gold standard} for influential robot behaviors.
The policies obtained from this framework optimize the robot's expected performance while reasoning over short- and long-term human dynamics.
However, computing exact or near-exact solutions to the MOMDP may not always be feasible.
As the dimension of the continuous states $x \in \mathcal{X}$ and actions $a_\mathcal{R} \in \mathcal{A}_\mathcal{R}$ increase, it becomes intractable to identify $\pi_\mathcal{R}$ with state-of-the-art MOMDP solvers.

A fundamental advantage to our unifying framework is that it provides a principled starting point for influence.
When the exact solution is infeasible, we can i) derive approximations that maintain influence while being more computationally efficient.
We can also ii) fairly compare different methods as simplifications of our overarching MOMDP framework.
For example --- instead of maintaining a belief distribution --- the robot can approximate the MOMDP by assuming a point estimate over $z$ and $\phi$.
Alternatively, the robot might assume that $z$ and $\phi$ will be revealed at a later interaction, and so the robot does not need to take information gathering actions.

To demonstrate the value of our unifying framework, below we show how the current approaches for influence described in Section~\ref{sec:existing} are approximations of our MOMDP.

\subsubsection{To Game-Theoretic Approaches}

Here we derive the game-theoretic methods from \eq{E2} and \eq{E3} (also tested in Section~\ref{sec:E2}) as approximations of our unified approach.
First, we slightly modify the human's observations: by default, our unified approach assumes that the human can only observe the robot's previous actions.
But for this approximation we extend the observations so that the human has access to the robot's next $H$ actions.
Let \eq{I3} in our human model be:
\begin{equation} \label{eq:I7}
    z^{t+1} = g_s(s^t, \mathbf{a}_\mathcal{R}, a_\mathcal{H}^t, z^t, \phi^t) = \mathbf{a}_\mathcal{R}
\end{equation}
so that the human's short term dynamics are a prediction $\mathbf{a}_\mathcal{R}$ of what the robot will do.
For this simplification the robot does not maintain any long-term dynamics, and $\phi = \emptyset$.

Because the robot also has access to its future actions (i.e., the robot knows what actions it is planning to take) here $z$ is fully observable.
This means that the augmented state $x = (s, z, \phi)$ is observable, and accordingly the MOMDP reduces to a Markov decision process (MDP). 
Within this MDP where the robot searches for actions $\mathbf{a}_\mathcal{R}$ that will maximize its cumulative reward $R_\mathcal{R}$.
Simultaneously, the human observes $z = \mathbf{a}_\mathcal{R}$ and responds with actions $\mathbf{a}_\mathcal{H}$ that maximize their cumulative reward $R_\mathcal{H}$.
This exactly corresponds to the nested optimization problem outlined in \eq{E2} and \eq{E3}.

\subsubsection{To Latent Representation Approaches}

Next, we derive the latent representation methods from \eq{E2} and \eq{E4} as a simplification of our unified MOMDP framework.
As before, we begin by ignoring the long-term human dynamics. 
Here the human follows some rules $\phi$ when responding to the robot, but these rules do not change over time; hence, \eq{I4} reduces to a constant value $\phi^{t+1} = \phi^t$.
The key step for this derivation is to approximate our unified MOMDP as a QMPD \citep{littman1995learning}.
Under the QMDP approximation the robot assumes that it will uncover the augmented state $(s, z, \phi)$ at the next timestep.
Accordingly, the robot does not need to take information gathering actions, and the QMDP simplifies into two subproblems: i) estimating the augmented state $x$ and ii) finding optimal actions given the current estimate of $x$.

To perform step i), existing approaches such as \cite{xie2021learning, bajcsy2024learning, parekh2023learning} directly observe system state $s$ and obtain a point estimate of $z$ and $\phi$.
Hence, the robot believes the augmented state is $\hat{x} = (s, \hat{z}, \hat{\phi})$.
Treating this point estimate as the current state, in step ii) the robot solves an MDP to obtain the policy $\pi_\mathcal{R}$ that maximizes the cumulative reward starting at $\hat{x}$ and evolving according to \eq{I6}.
Unwinding our definitions, the optimal actions for this MDP correspond to the solutions of \eq{E2} and \eq{E4}.

Interestingly, we notice that both existing approaches for influence simplify our unified framework by ignoring the long-term human dynamics $g_l$.
This is consistent with our experimental findings from Section~\ref{sec:E2}, and suggests that these methods fall short over repeated interactions because they fail to explicitly account for how the human will adapt to the influential robot over time.

%% file: 6_simulations.tex
\section{Simulating Influential Robots}\label{sec:sims}

Here we perform two experiments with simulated humans under controlled conditions.
The purpose of these simulations is to compare our unified framework from Section~\ref{sec:U1} with the two state-of-the-art simplifications discussed in Section~\ref{sec:U2}.
First, we implement our \textbf{Unified} and the game-theoretic \textbf{Stackelberg} approach in a simulated highway environment (see Section~\ref{sec:sim1}).
Second, we test the performance of our \textbf{Unified} framework against a \textbf{Latent} representation method within pursuit-evasion, driving, and robot manipulation environments (see Section~\ref{sec:sim2}).
Across both experiments we find that robots using our proposed formalism are more successful in influencing simulated humans, particularly over long-term interactions where the simulated agents completed the same tasks multiple times.
Code for our simulations and experiments is available here: \url{https://github.com/VT-Collab/influence}

\subsection{Comparing to Game-Theoretic Approaches} \label{sec:sim1}

\begin{figure*}[t]
	\begin{center}
        \includegraphics[width=2\columnwidth]{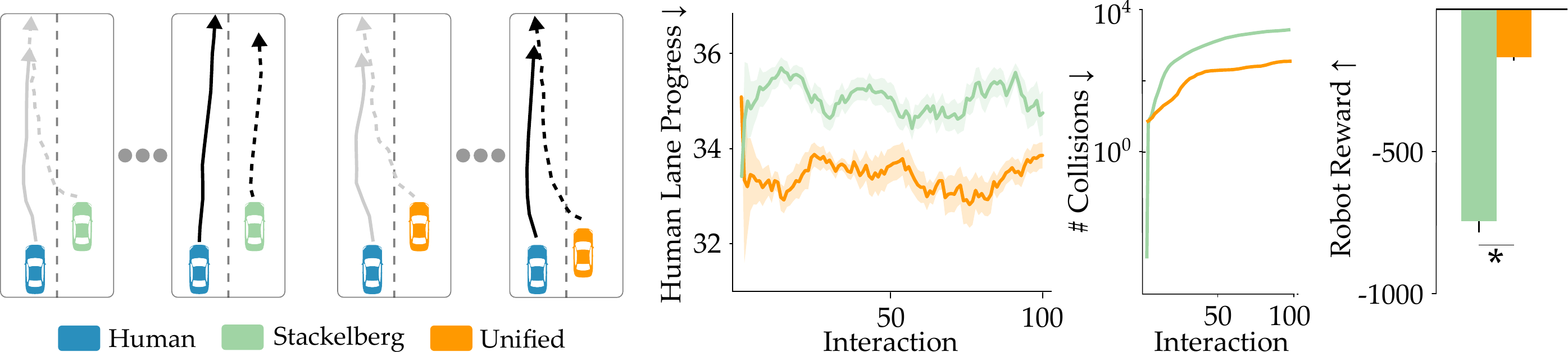}
		\caption{Experiment from Section~\ref{sec:sim1}. (Left) Simulated humans drive alongside robots using \textbf{Unified} and \textbf{Stackelberg} methods. The robot car takes actions to slow down the human without crashing. We show examples of the trajectories that the cars followed initially (i.e., first interactions) vs later on (i.e., final interactions). (Right) We plot average lane progress and total number of collisions per interaction. A higher lane progress indicates that the robot is unable to successfully influence the human to slow down. The shaded regions and error bars indicate standard error. We compare robot reward averaged across all interactions for each method. Fewer collisions and a higher reward are indicative of better robot performance. Asterisks $*$ denote statistical significance ($p < .05$).}
		\label{fig:sim1}
	\end{center}
    \vspace{-2.0em}
\end{figure*}

Our first simulation can be seen as a continuation of the user study from Section~\ref{sec:E2}.
Here simulated humans drive alongside an autonomous car on a highway environment (see \fig{sim1}); as before, the autonomous car's objective is to safely slow down the human driver.
The simulated human attempts to get around the autonomous car and maximize their lane progress by optimizing \eq{U2}.
Both agents have continuous state-action spaces.

\p{Simulated Human} To simulate the human, we implemented an agent that treats the interaction as a Stackelberg game: this game-theoretic human approximates the interaction as a turn-based sequence and takes actions according to \eq{E2} and \eq{E3}.
The human initially plays second (i.e., the human responds to the robot), but based on the behavior of the autonomous car the simulated human can shift between playing first or second in the Stackelberg game.
When playing first the human takes a more proactive role --- i.e., the human expects the robot to respond to its actions.
If the autonomous car merges in front of the human in more than $3$ of the past $6$ interactions, the human plays first (i.e., the human becomes more aggressive). 
Alternatively, if the human and autonomous car crash, the human reverts to playing second (i.e., the human becomes more defensive).
The autonomous car does not know how humans will behave \textit{a priori}.
Simulations were performed in Julia \citep{egorov2017pomdps} based on the CARLO environment \citep{cao2020reinforcement}.

\p{Independent Variables}
We compared two algorithms for influencing the simulated human.
As a baseline we used \textbf{Stackelberg}, the game-theoretic approximation discussed in Section~\ref{sec:U2} and previously implemented in works such as \cite{sadigh2016planning, lazar2018maximizing, fisac2019hierarchical, schwarting2019social}.
We compared this baseline to our \textbf{Unified} framework.
As a reminder, our approach is based on a MOMDP formulation: to solve this MOMDP for a tractable approximation of the influential robot policy, we used the POMCPOW algorithm \citep{sunberg2018online}.

\p{Dependent Measures} 
We simulated every human for $100$ interactions. 
Each individual interaction lasted $T = 120$ timesteps. 
The initial state of the cars were chosen at random, but with the condition that the autonomous car had an initial $y$-value greater than the simulated car (i.e., the robot started in front of the human).
We measured total lane progress for both cars (in pixels), total of number of collisions between the cars, and robot's total reward using \eq{U1}.

\p{Results} 
Our results from the first simulation are presented in \fig{sim1}. 
On the left we show qualitative examples of the trajectories that the cars followed initially (within the first $10$ interactions) vs. later on (within the final $10$ interactions). 
In the middle we plot average lane progress and the total number of collisions per interaction. 
Finally, on the right we compare the robot's reward averaged across all interactions for each method. 
To analyze these results we performed paired $t$-tests and found that with \textbf{Unified} the human made significantly less lane progress by the end of the interaction ($t(99) = -10.73, p < 0.001$), the human and autonomous car had fewer collisions ($t(99) = -16.01, p < 0.001$), and the autonomous car achieved a higher average reward ($t(99) = 14.42, p < 0.001$).

Overall, these experimental results are aligned with our theoretical analysis.
\textbf{Stackelberg} is a simplification of \textbf{Unified} that does not take into consideration how the human's behavior might change over repeated interactions.
Here the human shifted their role within the Stackelberg game: switching from passively playing second to more proactively playing first.
With \textbf{Stackelberg} the autonomous car could not predict or account for these changes, resulting in failed influence and increased collisions.
But with our \textbf{Unified} framework the autonomous car could reason over how the human's responses would change (i.e., the robot modeled $\phi$ as playing first or playing second).
The \textbf{Unified} robot leveraged these long-term dynamics to maintain safe influence over $100$ interactions.

\subsection{Comparing to Latent Representation Approaches} \label{sec:sim2}

\begin{figure*}
    \centering
    \includegraphics[width=2\columnwidth]{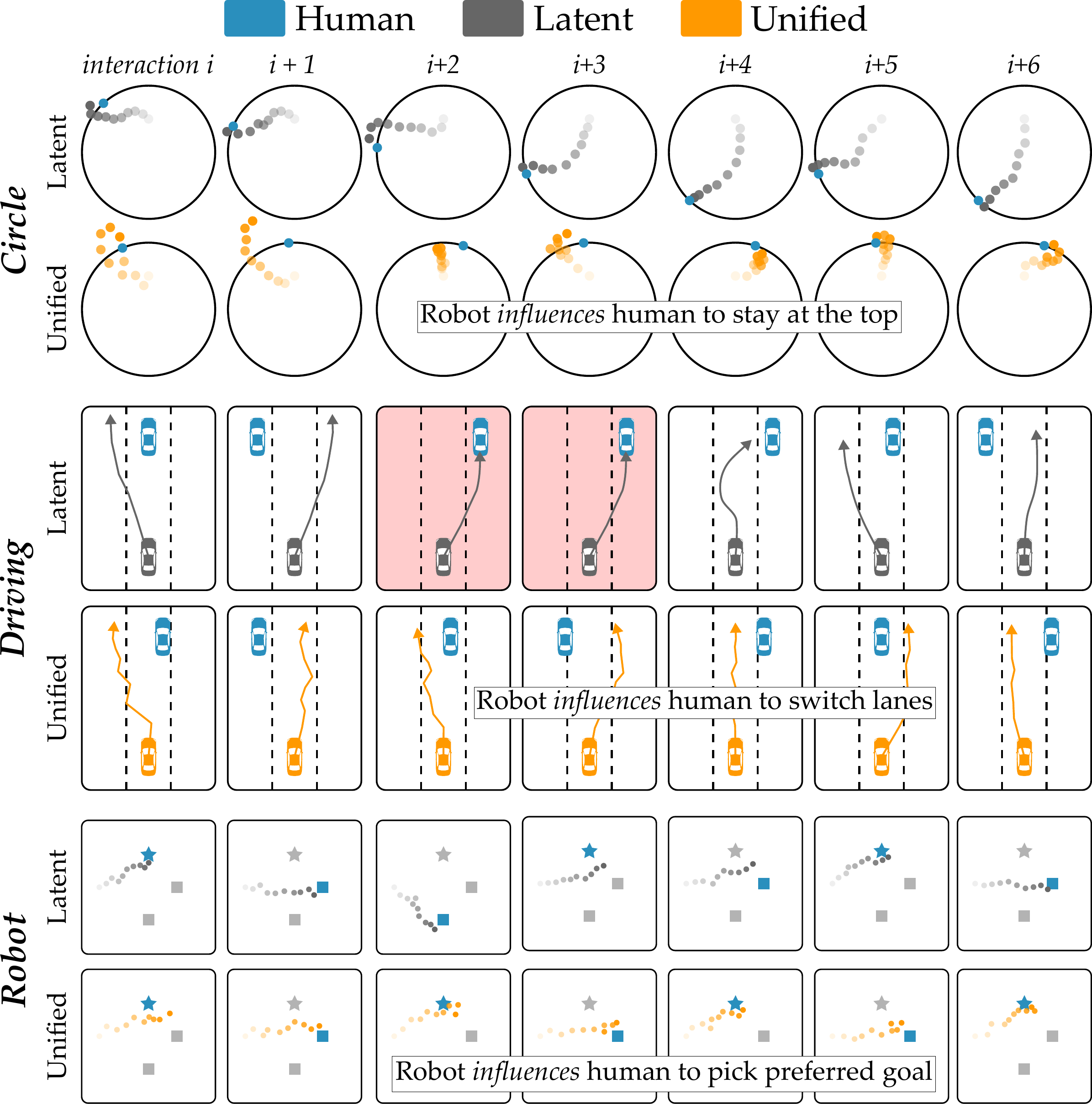}
    \caption{Example interactions from Section~\ref{sec:sim2} between the robot and simulated human in the Circle (Top), Driving (Middle), and Robot (Bottom) environments. The first row for each environment shows interactions with a robot using the \textbf{Latent} method, while the second row shows interactions with a robot using the \textbf{Unified} method. In Circle, \textbf{Unified} is able to successfully influence the human by trapping them at the top of the circle while \textbf{Latent} is not. In Driving, \textbf{Latent} leads to higher number of collisions since it is not able to influence the human to stay out of the robot's lane. Finally, in the Robot environment while \textbf{Latent} cycles through each of the three goals, \textbf{Unified} influences the human to frequently pick the robot's preferred goal (shown with a star).}
    \label{fig:enter-label}
    \vspace{-1.7em}
\end{figure*}

In our second simulation we compare the performance of our \textbf{Unified} formalism against a state-of-the-art \textbf{Latent} representation \citep{parekh2023learning}.
This \textbf{Latent} algorithm is consistent with the human model in \eq{E4}: the robot maintains a low-dimensional representation of the human's strategy $z$, and learns how to update this strategy based on the human's behavior from repeated interactions.
More formally, the robot learns latent representations $z \in \mathcal{Z}$ and short-term dynamics $g_s$ directly from interaction data.
We test how \textbf{Latent} and \textbf{Unified} influence simulated humans across three environments: \textit{Circle}, \textit{Driving}, and \textit{Robot}.
In each environment the simulated human has different rules governing how they update their short-term and long-term behavior in response to the robot's actions.
We detail these rules and the environments below.

\p{Simulated Environments \& Humans}
Our three environments for the second simulation are shown in \fig{enter-label}.
We selected these environments to be consistent with prior work \citep{parekh2023learning}.
All environments are composed of continuous state-action spaces.
Within each environment the robot repeatedly interacts with a simulated human $100$ times; each interaction has $T=10$ timesteps.
We repeat this process $50$ times (to simulate $50$ different humans) and report the averaged results.
Overall, the robot's objective is to maintain its influence across long-term interaction.

\p{Circle} 
The \textit{Circle} environment is an instance of pursuit-evasion games \citep{vidal2002probabilistic} with two-dimensional states and actions. 
The robot agent (i.e., the pursuer) tries to reach the simulated human agent (i.e., the evader).
To avoid the robot the human moves along the circumference of the circle. 
Here $z$ encodes where the human hides from the robot along the circle, and $\phi$ governs how the human moves along the circle to avoid the robot.
For example, some human evaders move away from the robot's previous position.
The robot's reward is its negative distance from the human.

\p{Driving} 
In \textit{Driving} an autonomous car is trying to pass a human vehicle.
At every interaction the human starts out in front of the autonomous car, and changes lanes as the autonomous car attempts to pass.
Here $z$ encodes the lane that the human merges into, and $\phi$ determines how the human selects that lane.
For instance, the some simulated humans merge into the lane where the autonomous car passed at the previous interaction.
The robot is rewarded for passing the human, and penalized for crashing with the human.
Similar to the \textit{Circle} environment, the robot does not know which lane the human will select; to safely pass the human the robot must anticipate and influence the driver's behavior.

\p{Robot} 
Finally, in \textit{Robot} the autonomous agent reaches for goals within a shared workspace.
The robot is rewarded if it reaches for the same goal as the simulated human (i.e., if both agents move to the far left object).
The robot’s action space is its $3\text{-DoF}$ end-effector velocity, the latent representation $z$ encodes the target the human wants to reach, and $\phi$ determines how the human updates their target object.
For instance, at interaction $i+1$ some humans select the object to the left of its target at the previous interaction $i$.

\p{Implementation} 
To ensure a fair comparison between the baseline \textbf{Latent} and our proposed \textbf{Unified} approach, we provided the system state $s = (s_\mathcal{R}, s_\mathcal{H})$ to both algorithms.
This system state included the position of the robot and the human.
Here providing the robot with $s$ alleviated the need to learn the latent strategy $z$, since the human's current strategy was often captured by their state (i.e., $z$ in \textit{Circle} encodes the human's location).
However, we emphasize that the robot does not know how the human will update this $z$ between interactions, i.e., $\phi$ is still a latent parameter that must be learned.
Similar to Section~\ref{sec:sim1}, we implemented our MOMDP framework using the Julia programming language \citep{egorov2017pomdps} and the POMCPOW approximation \citep{sunberg2018online}.

\begin{figure}[t]
	\begin{center}
        \includegraphics[width=1\columnwidth]{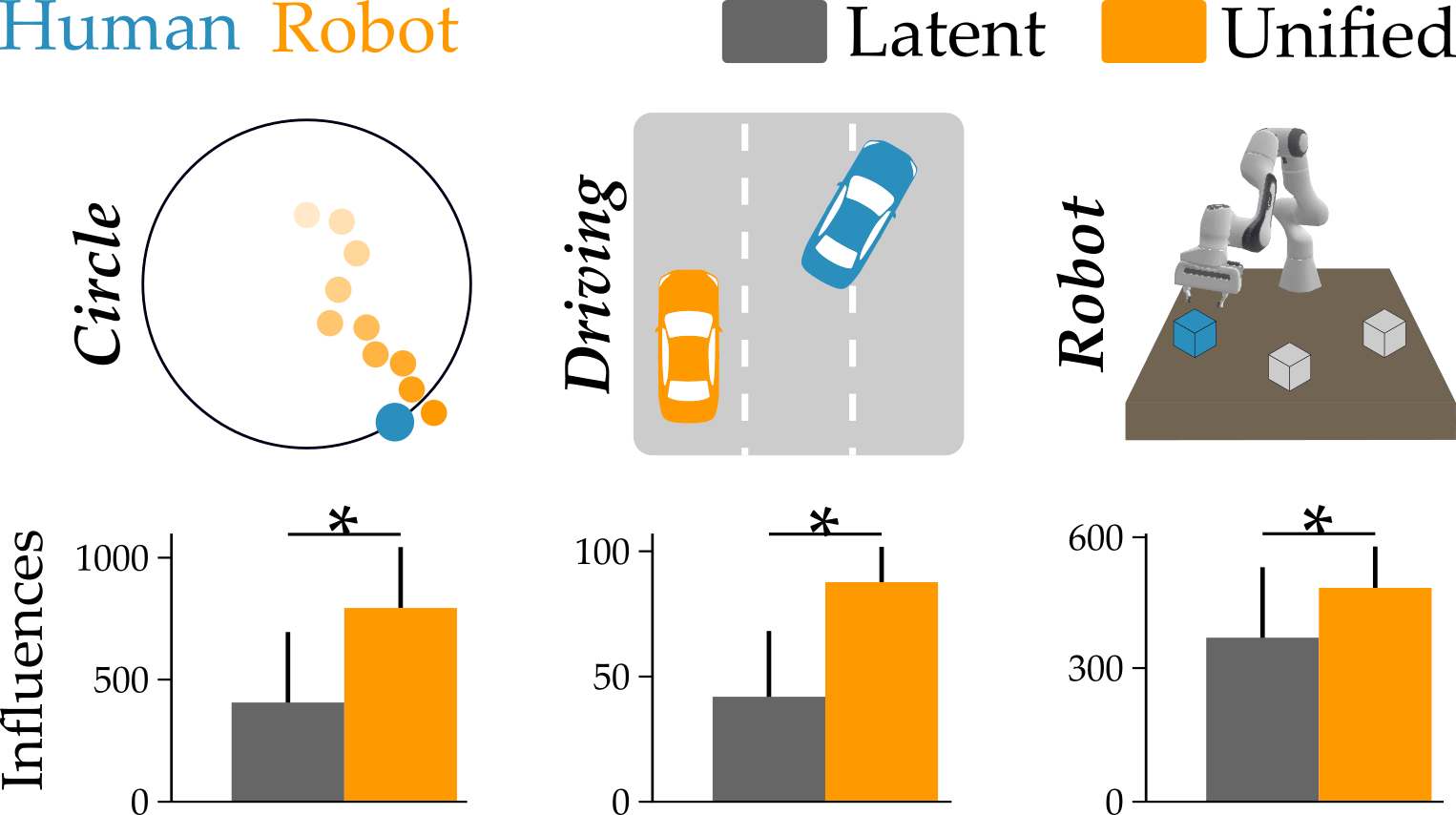}
		\caption{Experimental results from Section~\ref{sec:sim2} comparing \textbf{Latent} and \textbf{Unified} with simulated humans. The results are shown for the three environments Circle (Left), Driving (Middle), and Robot (Right). Plots show the number of interactions where the robot manages to successfully influence the human. Asterisks $*$ denote statistical significance ($p < .05$).}
		\label{fig:sim2}
	\end{center}
    \vspace{-2.0em}
\end{figure}

\p{Results} 
Our results from the second simulation are shown in \fig{sim2}. 
In the middle we provide quantitative examples of how each method interacted with simulated humans, and on the right we plot the number of times the robot successfully influenced the human.
The definition of successful influence varied based on the robot's reward.
For \textit{Circle} this was the number of times the robot pursuer trapped the human evader; for \textit{Driving} this was the number of times the human driver moved out of the autonomous car's way; and for \textit{Robot} this was the number of times the human selected the object closest to the robot.
Robots that applied our \textbf{Unified} framework were able to influence simulated human more frequently over long-term interactions than the \textbf{Latent} baseline.
Paired $t$-tests revealed that the number of successful influences for \textbf{Unified} was significantly higher than \textbf{Latent} in \textit{Circle} ($t(49) = -6.543, p < .001$), \textit{Driving} ($t(49) = -10.933, p < .001$), and \textit{Robot} ($t(49) = -4.411, p < .001$).

Overall, we again found that our experimental results aligned with our theoretical analysis.
The \textbf{Latent} simplification was able to learn the human's latent representation $z$ and the patterns the human leveraged to update $z$ in the short-term.
However, when humans changed the rules that underpinned $z$ (i.e., when the long-term dynamics caused $\phi$ to shift) \textbf{Latent} was confused, and the robot needed to re-learn the human's latent strategy and response pattern.
By contrast, \textbf{Unified} constantly reasoned about both short-term and long-term dynamics, ultimately learning not just how the human would update $z$ for the next interaction, but also how the robot's actions could impact $\phi$.
The robot's ability with \textbf{Unified} to account for long-term impacts enabled the robot to more frequently influence the simulated human.

%% file: 7_user-studies.tex
\begin{figure*}[t]
    \vspace{0.42em}
	\begin{center}
		\includegraphics[width=2\columnwidth]{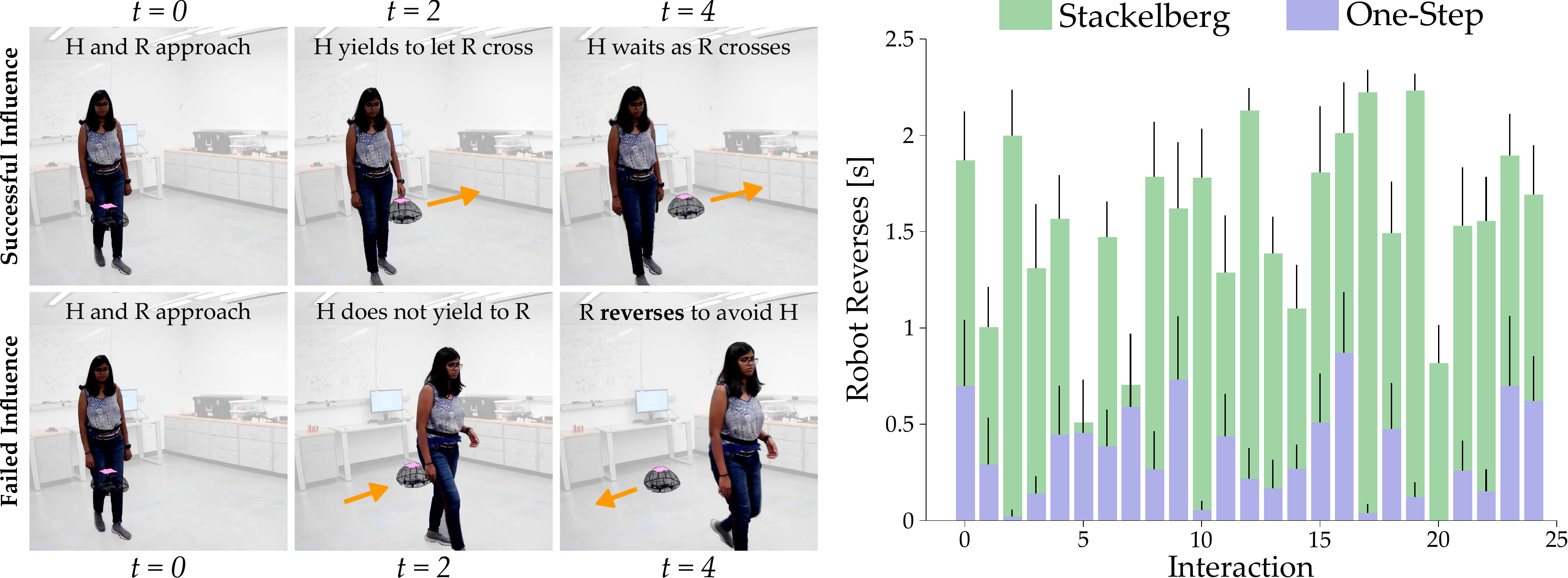}
		\caption{User study from Section~\ref{sec:user1} where in-person participants repeatedly cross paths with a drone. (Left) When the drone and human intersect, the drone tries to influence humans to yield so that it can cross first. If this influence fails, the robot temporarily reverses direction to avoid a collision. (Right) We plot the amount of time the robot reverses direction across $25$ repeated interactions. The \textbf{Stackelberg} drone solves \eq{E2} and \eq{E3} to try and influence the human; this robot always yielded once the human was within a specific radius. By contrast, the \textbf{One-Step} drone used a simplification of the unified framework presented in \eq{user1} to make its actions less predictable. This drone would occasionally yield at a larger radius or a smaller radius as compared to the \textbf{Stackelberg} drone. We found that this unpredictability reduced the amount of time the robot had to reverse and increased the number of successful influences. Error bars show standard error.}
		\label{fig:us2}
	\end{center}
	\vspace{-2.0em}
\end{figure*}

\section{Influencing Users in the Long-Term} \label{sec:user}

In Section~\ref{sec:sims} we demonstrated how our unified formulation can influence simulated humans over repeated interactions.
Equipped with this formulation, we now return to our core challenge: controlling how robots influence actual users over long-term interaction.
In Section~\ref{sec:existing} we highlighted that existing approaches fail to regulate this long-term influence --- within a span of $12$ interactions, humans quickly learned to disregard the robot's actions.
Now we return to similar experimental setups, and seek to influence participants over the span of $\mathbf{25+}$ \textbf{interactions}.
We demonstrate two contributions of our unified approach.
First, in Section~\ref{sec:user1} we show how designers can leverage our unified framework to generate principled but tractable simplifications.
We make a greedy approximation (i.e., a one-step look-ahead) to arrive at a new algorithm for influencing humans, and test how that method influences users during human-drone interactions.
Second, in Section~\ref{sec:user2} we leverage particle-based solvers to reach a near-exact solution to our unified framework, and compare that approach to state-of-the-art baselines.
Here users drive a simulated vehicle alongside an autonomous car: we again test whether the robot is able to consistency regulate how aggressively the human drives.
Videos of our user studies are available online at: \url{https://youtu.be/nPekTUfUEbo}.

\subsection{Interacting with an Aerial Drone} \label{sec:user1}

In this experiment we test how our unified framework can be leveraged to reach principled but tractable simplifications.
In general, computing near-exact solutions to the MOMDP may not always be feasible --- but designers can introduce approximations into this framework to derive real-time methods.
Game-theoretic and latent approaches to influence are two examples of these approximations.
Here we generate another novel approximation: the \textbf{One-Step} method.
We then compare that approach to the \textbf{Stackelberg} baseline in a user study where participants are walking near an aerial drone.
The humans and drone have orthogonal goals, and often need to intersect each other's path.
At these intersections the drone takes actions to attempt to influence the humans to yield (so that the drone passes first and completes its task more efficiently).
We measure how successfully the drone influences the human over $25$ repeated interactions.

\p{One-Step Approximation}
On one hand, fully solving the MOMDP requires the robot to reason about all possible evolutions of its belief.  
On the other hand, if the robot does not consider how its belief can change, then it will be unable to reason about how the person could be influenced and thus fail to maintain control. 
Our \textbf{One-Step} approximation to the MOMDP in Section~\ref{sec:unified} attempts to strike a balance between these two extremes by having the robot reason about one belief update it will perform after the current timestep.
Let $b$ be the robot's belief over the augmented state $x = (s, z, \phi)$, where $z$ and $\phi$ are both unknown parameters of the human model.
We instantiate $z$ as the human's reward function, such that $z = r_\mathcal{H}$, and we instantiate $\phi$ as the human's estimate of the robot's reward, such that $\phi = b(r_\mathcal{R})$.
Within this instantiation the human is trying to determine what the robot is optimizing for: as the human uncovers $r_\mathcal{R}$, the human can accurately anticipate or ignore the robot's influential actions.

To tractably solve this instantiation for the optimal robot policy $\pi_\mathcal{R}$, we assume that the robot will fully observe $x$ \textit{two timesteps in the future}.
This is an improvement over the QMDP approximation (where the robot assumes it will uncover $x$ at the \textit{next} timestep).
As a result, our \textbf{One-Step} look-ahead greedily reasons over how its belief $b$ will evolve, and incorporates that belief within its objective.
Using this one-step simplification we reach the modified reward function \citep{brooks2019balanced}:
\begin{multline} \label{eq:user1}
    R_{\mathcal{R}}(s^t, \mathbf{a}_{\mathcal{R}}, \mathbf{a}_{\mathcal{H}}) = \lambda \mathcal{H}(\phi^{t+1}) + \\ \sum_{s \in \xi(s^t, \mathbf{a}_{\mathcal{R}}, \mathbf{a}_{\mathcal{H}})} r_\mathcal{R}(s, \theta_\mathcal{R})
\end{multline}
where $\mathcal{H}$ is the Shannon entropy over the distribution $\phi$ at one timestep in the future, and $\lambda \geq 0$ is hyperparameter set by the designer.
Comparing \eq{user1} to \eq{E1}, we see that the key difference is the inclusion of the entropy term.
Maximizing this modified reward causes the robot to take actions \textit{now} that will increase the human's uncertainty over the robot's reward at the \textit{next} timestep.
Put another way, the robot proactively attempts to modify $\phi$ so that the human cannot uncover what the robot optimizing for or predict how the robot will behave.
We expect that this \textbf{One-Step} simplification will outperform existing methods such as \textbf{Stackelberg} because i) the robot is reasoning over how $\phi$ evolves within long-term human dynamics and ii) the robot attempts to control that evolution so that the robot can maintain influence.

\p{Experimental Setup} 
To put this simplification generated by our unified approach to the test, we created a task where participants interacted with an aerial drone (see \fig{us2}).
The drone and human had similar tasks at opposite sides of the room.
Participants walked back and forth across the room to pick up blocks and build a tower; each time the human started to cross, the drone moved orthogonally to monitor the blocks at two other locations.
This resulted in repeated intersections between the drone and the human.
Both agents had to determine how to navigate these intersections: e.g., the drone might move backwards to yield to the human, or the human might wait for the drone to cross.
We tracked the position of the drone using ceiling-mounted cameras, and humans wore an HTC Vive Tracker around their waist for real-time position measurements.

\p{Independent Variables} 
We compared two robot controllers: the \textbf{Stackelberg} baseline introduced in Section~\ref{sec:existing} and the \textbf{One-Step} simplification of our unified framework.
The robot was rewarded for crossing the room as quickly as possible while avoiding collisions with the human: to maximize its speed, the robot tried to influence humans to yield. 
The drone selected actions in real-time using the following reward functions:
\begin{equation} \label{eq:U2_1}
    r_\mathcal{R}(s, \theta_\mathcal{R}) = \dot{s}_{\mathcal{R}} - 10 \cdot \mathbb{1}\{\text{collision in }s\}
\end{equation}
\begin{equation} \label{eq:U2_2}
    r_\mathcal{H}(s, \theta_\mathcal{H}) = \dot{s}_{\mathcal{H}} - 100 \cdot \mathbb{1}\{\text{collision in }s\}
\end{equation}
where $\dot{s}_{\mathcal{R}} \subset s$ is the robot's forward velocity and $\dot{s}_{\mathcal{H}} \subset s$ is the human's velocity. 
We recognize that actual participants may not have followed this human reward function: $r_\mathcal{H}$ is used purely by the robot to calculate the \textbf{Stackelberg} human model in \eq{E3}. 

Negative values for $\dot{s}_{\mathcal{R}}$ indicate that the drone is reversing direction and yielding to the human.
Recall that the \textbf{One-Step} robot takes actions that make the human uncertain about what the robot is optimizing for (i.e., whether the robot is trying to avoid collisions). 
In practice, this caused the \textbf{One-Step} drone to randomly switch between crossing aggressively (only yielding if the human was within a small radius) and defensively (yielding if the human was anywhere within a larger radius).

\p{Dependent Measures} 
A robot that maintains the right-of-way will always have a positive $\dot{s}_{\mathcal{R}}$.
However, if the human insists on going first, then the robot must back off and give the participant space. 
To measure influence, we therefore recorded the amount of time the robot \textit{reversed} during each interaction. 
Lower values correspond to higher influence (i.e., a robot that reverses direction the fewest number of times is effectively influencing the human to yield).

\p{Participants} 
We recruited $11$ participants from the Virginia Tech community ($10$ male, ages $22.1 \pm 3.1$ years). 
These participants provided informed written consent under IRB \#$20$-$755$. 
We recognized that people may hesitate to walk close to a flying drone; before starting the experiment, we accordingly demonstrated the task and drone behaviors to help users become familiar and comfortable walking near the drone.
We leveraged a within-subjects design: all participants interacted with a \textbf{One-Step} robot $25$ times and a \textbf{Stackelberg} robot $25$ times. 
The order of presentation was balanced across users, so that half started with \textbf{One-Step} and the other half started with \textbf{Stackelberg}.

\p{Hypothesis} We hypothesized that:
\begin{quote}
\p{H2} \textit{A drone that approximates our unified framework while reasoning about long-term influence will cause the human to yield more consistently than the game-theoretic baseline.}
\end{quote}

\p{Results} 
Our results are summarized in \fig{us2}. 
Paired $t$-tests reveal that the \textbf{Stackelberg} robot spent significantly more time backing-up and yielding to the human as compared to \textbf{One-Step} ($t(327) = 13.02, p < .001$). 
We also noticed that --- as participants became more familiar with the \textbf{Stackelberg} robot --- they insisted on going first more frequently (perhaps because they were able to predict when this robot would yield). 
As a result, the final $8$ interactions with \textbf{Stackelberg} had a higher average reverse time than the first $8$ interactions, although this difference was not statistically significant ($t(10) = -1.45, p = .09$). 
Overall, the findings from this study supported \textbf{H2}: the \textbf{One-Step} robot consistently influenced the $11$ participants to yield the right-of-way over $25$ repeated interactions.
This user study demonstrates that we can leverage our unified framework to generate novel simplifications (such as \textbf{One-Step}), and by carefully designing these simplifications we reach robots that effectively influence humans over long-term interaction.

\subsection{Driving Alongside an Autonomous car} \label{sec:user2}

\begin{figure*}[t]
	\begin{center}
        \includegraphics[width=2\columnwidth]{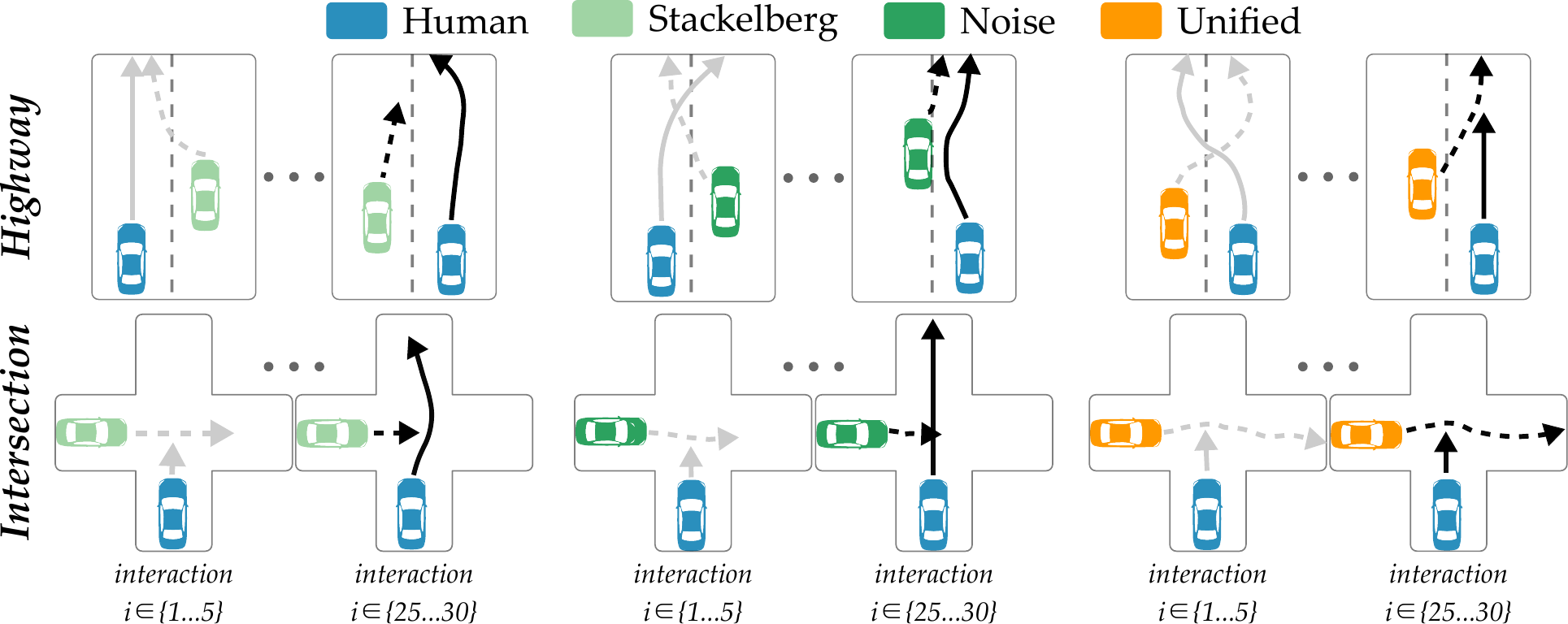}
		\caption{Example interactions from Section~\ref{sec:user2} between the robot and a user in Highway (Top) and Intersection (Bottom). The interactions are shown for robots using the three frameworks, \textbf{Stackelberg}, \textbf{Noise}, and \textbf{Unified}. We compare how influential the robot was in the beginning of the experiment versus at the end. All three frameworks enable the robot to successfully influence the user at the start. In Highway, the robot influences the user to slow down, reducing their lane progress. In Intersection, the robot influences the user to yield. Over long-term, repeated interactions with $N=20$ users, only the \textbf{Unified} framework is able to maintain influence --- slowing down the user in Highway or crossing in front of the user in Intersection.}
		\label{fig:us3_quantitative}
	\end{center}
    \vspace{-2.0em}
\end{figure*}

In our first user study (Section~\ref{sec:user1}) we demonstrated how our unified framework can be leveraged to generate influential algorithms.
This makes sense for cases where the problem setting is high dimensional, and we cannot practically solve the MOMDP.
At the same time, we recognize that recent advances in optimization enable robots to reach near-exact solutions to the MOMDP in increasingly complex settings \citep{kurniawati2022partially}.
In this experiment we therefore put our unified formalism to the test, and compare \textbf{Unified} to two baselines.
The user study is performed in the same driving environment as in Sections~\ref{sec:E2} and \ref{sec:sim1}.
Participants drive a simulated vehicle alongside an autonomous car in both a highway and intersection environment.
The autonomous car attempts to regulate the speed of the human on the highway, and cross in front of the human at the intersection.
Importantly, this study focuses on long-term influence: participants interact with the autonomous car $30$ times in each environment.
Our results explore whether the unified formulation --- in its complete form --- regulates influence more effectively than state-of-the-art approximations.

\p{Experiment Setup}
We created driving environments consistent with both prior works on influence \citep{tian2023towards, sadigh2016planning, fisac2019hierarchical, schwarting2019social} and our preliminary experiments from Sections~\ref{sec:E2} and \ref{sec:sim1}.
Participants shared a road with an autonomous car in two settings: Highway and Intersection (see \fig{us3_quantitative}).
The vehicles had point-mass dynamics from the CARLO environment \citep{cao2020reinforcement}.
To control their car in real-time, users interacted with a Logitech G29 steering wheel and pedal.
The length of the interactions was fixed; the next interaction started immediately after the previous one ended.
As described in \eq{U2}, participants earned a score based on staying on the road, making lane progress, and avoiding collisions with the autonomous car.
This score was displayed to participants throughout the experiment to encourage safe and efficient driving.

\p{Independent Variables} 
We varied the autonomous car's control strategy with three levels.
To provide a baseline consistent with our previous studies, we included the game-theoretic \textbf{Stackelberg} algorithm.
Next, we tested a modified version of this Stackelberg approach that added zero-mean Gaussian noise to the robot's actions.
Under \textbf{Noise} the robot solves \eq{E2} and \eq{E3} to find its action trajectory $\mathbf{a}^*_{\mathcal{R}} = (a_{\mathcal{R}}^{*,0}, \hdots, a_{\mathcal{R}}^{*,T})$, and then we inject noise at each timestep:
\begin{equation} \label{eq:user2}
    a_{\mathcal{R}}^t = a_{\mathcal{R}}^{*,t} + \epsilon^t, \quad \epsilon^t \sim \mathcal{N}(0, \Sigma)
\end{equation}
Intuitively, we thought that \textbf{Noise} might improve \textbf{Stackelberg} because it makes the robot's actions less predictable, preventing the human from precisely anticipating the robot's influential behavior over repeated interactions.
Against the these baselines we compare our \textbf{Unified} formalism as described in Section~\ref{sec:U2}.
To tractably obtain $\pi_\mathcal{R}$ we again leveraged the POMCPOW approximation in Julia \citep{sunberg2018online}.
For each of the implemented controllers the robot reward was the same as \eq{U1}.

\p{Dependent Measures} 
For both environments we measured the human's lane progress, total number of collisions, and robot reward. 
A participant that speeds up and safely passes the autonomous car will increase their own score while lowering the robot's reward.
Here the autonomous car successfully influences the human towards defensive driving by (i) reducing the speed of the human on the \textit{Highway} or (ii) crossing the \textit{Intersection} in front of the human.

\p{Participants and Procedure} 
We advertised our experiment to the Virginia Tech community and recruited $20$ college-age participants ($16$ male, ages $23.75 \pm 2.79$ years). 
All the participants had experience with driving vehicles, and $16$ of the users had previously played racing video games. 
Participants provided informed written consent following university guidelines (IRB $\#23$-$784$). 

At the start of this user study participants answered a demographics survey. 
Next --- similar to the experimental procedure in Section \ref{sec:E2} --- participants practiced driving in each environment for roughly five minutes.
After this familiarization phase we began the recorded trials by following a within subjects design.
Every participant interacted with \textbf{Stackelberg}, \textbf{Noise}, and \textbf{Unified} $60$ times ($30$ per environment) so that individual users completed a total of $60 \times 3 = 180$ trials.
The order of the tasks and influential algorithms was randomized and balanced across participants.
After interacting with each algorithm the participants answered survey questions about their experience.
On a $1$-to-$7$ Likert scale, users indicated how much they agreed with the following prompts: the autonomous car's response was \textit{predictable}, and I \textit{changed my inputs} based on what the autonomous car was doing.

\p{Hypothesis} We hypothesized that:
\begin{quote}
\p{H3} \textit{Robots controlled with our unified framework will more effectively influence actual users over the long-term than current approaches.}
\end{quote}

\begin{figure*}[t]
	\begin{center}
        \includegraphics[width=2\columnwidth]{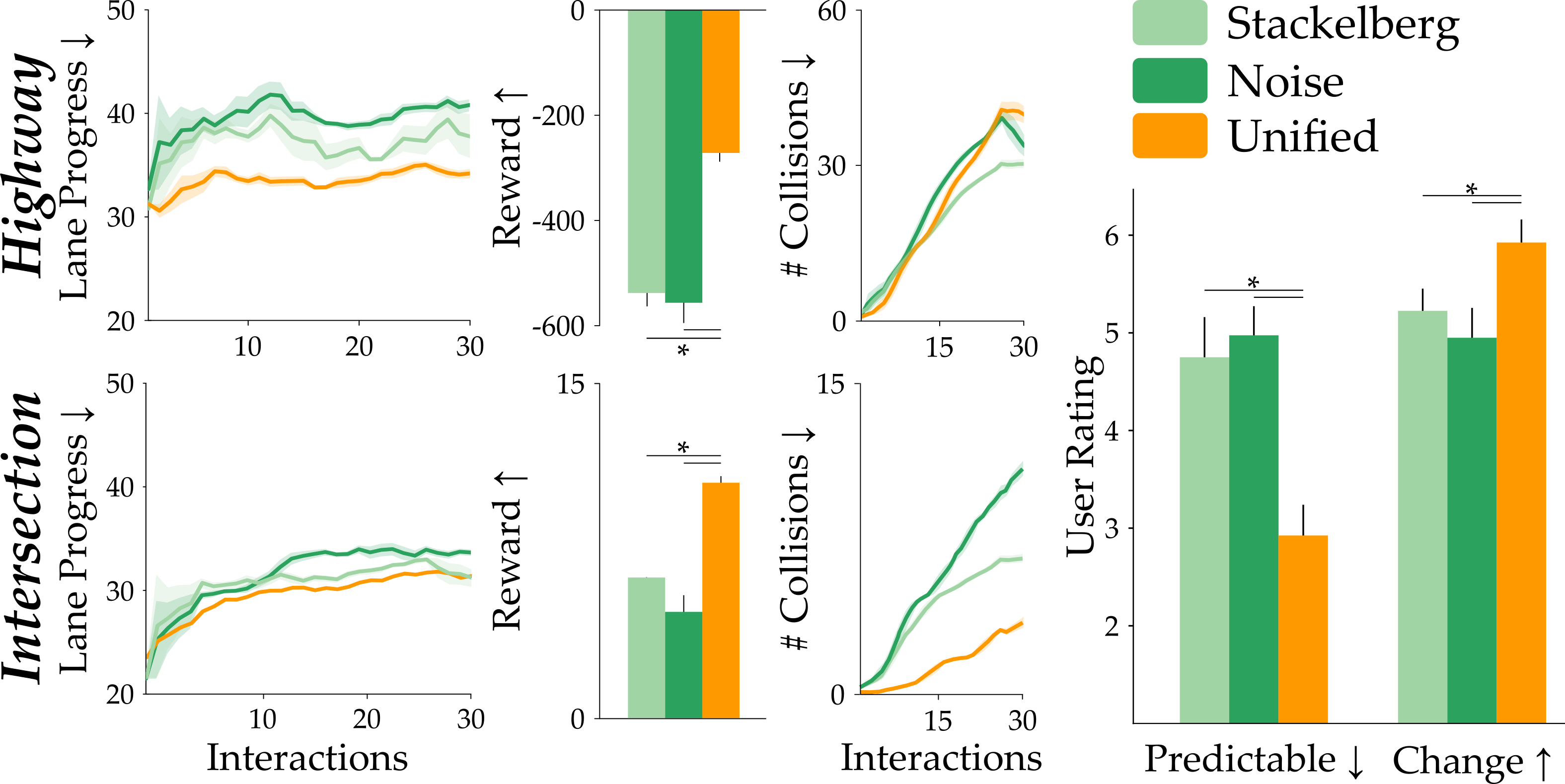}
		\caption{Objective and subjective results from our driving user study in Section~\ref{sec:user2}. (Left) Objective results for the highway and intersection tasks. The leftmost plots show the lane progress made by the human; a lower lane progress indicates that the robot is successfully able to influence the users to slow down. The plots in the second column compare the robot's reward across all the users. A higher reward suggests better robot performance. The plots in the third column display the number of collisions between the robot and the human in each interaction. The shaded region and the error bar in the plots show the standard error. (Right) Subjective results from Likert scale surveys. We plot the average user rating for \textit{Predictable}, i.e., whether participants perceived the robot's behavior to be easy to anticipate, and \textit{Change}, i.e. whether participants thought they changed their behavior in response to the robot's actions. A lower rating for \textit{Predictable} suggests that the robot could maintain its influence over humans who could not easily anticipate the robot's behavior. 
        In contrast, for \textit{Change} a higher rating suggests that the robot was able to influence the users.}
		\label{fig:us3}
	\end{center}
    \vspace{-2.0em}
\end{figure*}

\p{Results} 
Our average results across $20$ participants are summarized in Figures~\ref{fig:us3_quantitative} and \ref{fig:us3}.

In \fig{us3_quantitative} we show examples of the interaction trajectories for the two environments and three control algorithms.
These trajectories are presented in pairs: on the left of each pair are the participant and robot behaviors at the start of the experiment, and on the right are the participant and robot behaviors after $30$ repeated interactions.
In general, we observe that \textbf{Stackelberg}, \textbf{Noise}, and \textbf{Unified} robots are all able to initially influence the human to slow down or yield.
But as the human gains experience driving alongside the robot, with \textbf{Stackelberg} and \textbf{Noise} participants identify behaviors that avoid or mitigate the robot's influential actions.
By contrast, \textbf{Unified} is able to maintain influence --- consider \textit{Intersection}, where the \textbf{Unified} robot still crosses in front of the human driver at the end of $30$ trials.

The objective results from our final user study are presented in \fig{us3} (Left).
Here we plot the human's lane progress, the robot's reward, and the number of collisions.
Robots that influence participants should \textit{minimize} lane progress and collisions while \textit{maximizing} robot reward.
To analyze our results --- and examine whether each algorithm was able to influence humans --- we conducted repeated measures ANOVAs.
We found that algorithm type had a significant effect on lane progress (Highway: $F(1, 599) = 97.7, p < .001$, Intersection: $F(1, 599) = 25.9, p < .001$), robot reward (Highway: $F(1, 599) = 21.0, p < .001 $, Intersection: $F(1, 599) = 19.8, p < .001$) and collisions (Highway: $F(1, 599) = 8.6, p = .003$, Intersection: $F(1, 599) = 75.2 , p < .001$).
Bonferroni post hoc tests revealed that \textbf{Unified} caused the human to have less lane progress than \textbf{Stackelberg} ($p < .001$) or \textbf{Noise} ($p < .001)$ across Highway and Intersection.
Similarly, \textbf{Unified} increased the robot's reward over \textbf{Stackelberg} ($p < .001$) and \textbf{Unified} ($p < .001$).
For the Highway environment, the number of collisions was roughly equal for \textbf{Unified} and \textbf{Noise} ($p = 0.87$).
But in the Intersection environment \textbf{Unified} reduced the collisions as compared to both \textbf{Stackelberg} ($p < .001$) and \textbf{Noise} ($p < .001$).
A possible explanation for this difference in the number of collisions is that human drivers perceived the Intersection to be a more structured environment (with clear expectations for going first or second), while the Highway was less structured --- e.g., human drivers could weave between lanes repeatedly to try and pass the robot.
This may have led to more collisions across the board.

Finally, in \fig{us3} (Right) we plot the subjective results of our survey.
This survey had two items: \textit{Predictable} and \textit{Change}.
\textit{Predictable} indicated whether participants perceived the robot's behavior to be easy to anticipate, and \textit{Change} indicated whether participants thought they changed their behavior in response to the robot's actions.
Repeated measures ANOVAs show that users gave significantly different scores for \textit{Predictable} ($F(1, 19) = 11.0, p < .01$) and \textit{Change} ($F(1, 19) = 8.0, p < .05$) based on the type of robot they interacted with.
Post hoc tests suggest that with \textbf{Unified} participants found the robot to be \textit{harder to predict} than with \textbf{Stackelberg} ($p < .05$) or \textbf{Noise} ($p < .001$).
In addition, when interacting with \textbf{Unified} participants felt that they had to \textit{change their behaviors more} than with both \textbf{Stackelberg} ($p < .05$) and \textbf{Noise} ($p < .05$).
These subjective results match our previous findings for controlling influence.
When humans are able to easily anticipate what the robot will do, they can plan around those behaviors and mitigate influential actions.
By contrast, robots that effectively influence humans cause those humans to change their actions in response to the robot's behavior.

\p{Summary} 
Our final user study demonstrates that when robots solve our unified framework to select their policy, they are able to consistently influence humans across long-term interactions.
Specifically, the \textbf{Unified} approach guided $20$ participants to drive more defensively across $60$ total interactions within two driving environments.

%% file: 8_conclusion.tex
\section{Conclusion} \label{sec:conclusion}

Just as AI agents have affected human decision making, embodied agents will influence the way people interact.
In this paper we presented a unified framework to control how robots influence humans over long-term interaction.
We first demonstrated that existing approaches are often unable to regulate influence during repeated interactions; this failure occurs because existing approaches rely on a static human model, and assume humans respond to robot behaviors in a fixed way.
However, our experiments revealed that humans adapt to the robot --- so that behaviors which were originally influential are later avoided or ignored.

To enable robots to reason over how humans adapt to their actions, we next developed an optimization framework for long-term influence.
Our framework models humans as history-aware agents with short- and long-term dynamics, and then incorporates that human model into an augmented dynamical system.
We formally write this as a single-agent system (i.e., the robot is the actor), where parameters of the human's evolving model are unknown parts of the augmented state, and the robot gathers information about this state by observing the human's actions.
Our overall formulation is an instance of a mixed-observability Markov decision process (MOMDP): existing optimization tools can be leveraged to obtain near-optimal robot policies from our framework.
We refer to this formalism as \textit{unified} because we can derive existing approaches from our method.
For example, we show that state-of-the-art influential algorithms based on game theory and latent representations are actually simplifications of our MOMDP.

We tested our unified framework across simulations and user studies.
Our simulations compared the unified method against state-of-the-art simplifications.
These simplifications ignore the human's long-term dynamics --- so while they are able to influence the human at first, they struggle to maintain that influence as the human changes behavior.
Accordingly, the simulation results supported our theoretical analysis and indicated that solving the unified framework outperforms state-of-the-art approximations.
Next, we conducted two user studies where $N=11$ participants interacted with an aerial drone and $N=20$ participants drove a simulated vehicle alongside an autonomous car.
These user studies focused on long-term interactions: the participants worked with the robot for $25+$ trials.
Our results suggest that i) designers can apply new simplifications to our method to reach tractable but influential policies, and ii) our unified approach is able to successfully regulate how it influences humans over repeated interactions.

\p{Limitations}
This paper is a step towards robots that consider how their behaviors will affect human decision making in the short-term and long-term.
One limitation of this work is that our experiments focus on \textit{one robot interacting with one human}.
Although the theoretical framework we developed in Sections~\ref{sec:problem}--\ref{sec:unified} can extend to multiple users, it is not yet clear how this method will scale in practice.
Theoretically, interacting with several humans increases the dimension of the latent parameters $z = (z_1, \ldots, z_k)$ and $\phi = (\phi_1, \ldots, \phi_k)$.
Here $k$ is the number of humans, and $z_i$ and $\phi_i$ parameterize the robot's model of the $i$-th human.
Increasing the number of humans increases $k$, meaning that the latent vectors $z$ and $\phi$ gain dimensions.
This is a practical challenge for our approach: as discussed in Section~\ref{sec:U2}, solving the MOMDP becomes intractable as the belief space grows.
Designers can introduce simplifications to convert this multi-human MOMDP into a feasible optimization problem --- but without additional experiments, it is not year clear which simplifications are most appropriate.
Overall, we see the problem of one robot influencing multiple humans as an open question where future work can build upon our formalism.